\documentclass[10pt,twocolumn,letterpaper]{article}

\usepackage[pagenumbers]{cvpr} % To force page numbers, e.g. for an arXiv version

\usepackage{graphicx}
\usepackage{amsmath}
\usepackage{amssymb}
\usepackage{booktabs}
\usepackage{array}
\usepackage{multirow}
\usepackage[english]{babel}

\usepackage{xcolor}
\usepackage{caption}
\usepackage{lipsum}
\usepackage{enumerate}
\usepackage{subcaption}
\usepackage{blindtext}
\usepackage[T1]{fontenc}
\usepackage{footnote}
\usepackage[pagebackref,breaklinks,colorlinks]{hyperref}

\usepackage[capitalize]{cleveref}
\crefname{section}{Sec.}{Secs.}
\Crefname{section}{Section}{Sections}
\crefname{table}{Tab.}{Tabs.}
\Crefname{table}{Table}{Tables}

\makeatletter
\DeclareRobustCommand\onedot{\futurelet\@let@token\@onedot}
\def\@onedot{\ifx\@let@token.\else.\null\fi\xspace}
\def\eg{e.g\onedot} 
\def\ie{i.e\onedot}

\def\etal{et~al\onedot}

\makeatother

\definecolor{darkred}{rgb}{0.7,0.2,0.1}
\definecolor{darkgreen}{rgb}{0,0.7,0}
\definecolor{orange}{RGB}{255,127,0}
\definecolor{ourpurple}{RGB}{127,127,204}
\definecolor{palgreen}{RGB}{51,179,179}
\definecolor{magenta}{RGB}{199,21,133}

\def\cvprPaperID{9982} % *** Enter the CVPR Paper ID here
\def\confName{CVPR}
\def\confYear{2023}

\begin{document}


\title{ALTO: Alternating Latent Topologies for Implicit 3D Reconstruction}

% \maketitle
% \begin{figure*}[t]
%   \centering
% %   \fbox{\rule{0pt}{3.5in} \rule{\linewidth}{0pt}}
%   %\includegraphics[width=0.8\linewidth]{egfigure.eps}
%  \includegraphics[width=\textwidth]{figures/teaser.pdf}
%   \caption{\achuta{to add} }
% \end{figure*}
\author{Zhen Wang$^1$\thanks{Equal contribution.} \quad Shijie Zhou$^1$\footnotemark[1] \quad  Jeong Joon Park$^2$
\quad Despoina Paschalidou$^2$ \\ Suya You$^3$ \quad  
Gordon Wetzstein$^2$ \quad  Leonidas Guibas$^2$ \quad  Achuta Kadambi$^1$\\
\normalsize{$^1$University of California, Los Angeles \quad
$^2$Stanford University \quad
$^3$DEVCOM Army Research Laboratory}
% {\tt\small {zhenwang, shijiezhou}@ucla.edu}
% \and
% \\
% \\
}
\twocolumn[{%
\renewcommand\twocolumn[1][]{#1}%
\maketitle
    \captionsetup{type=figure}
    \vspace{-2mm}
    \includegraphics[width=\textwidth]{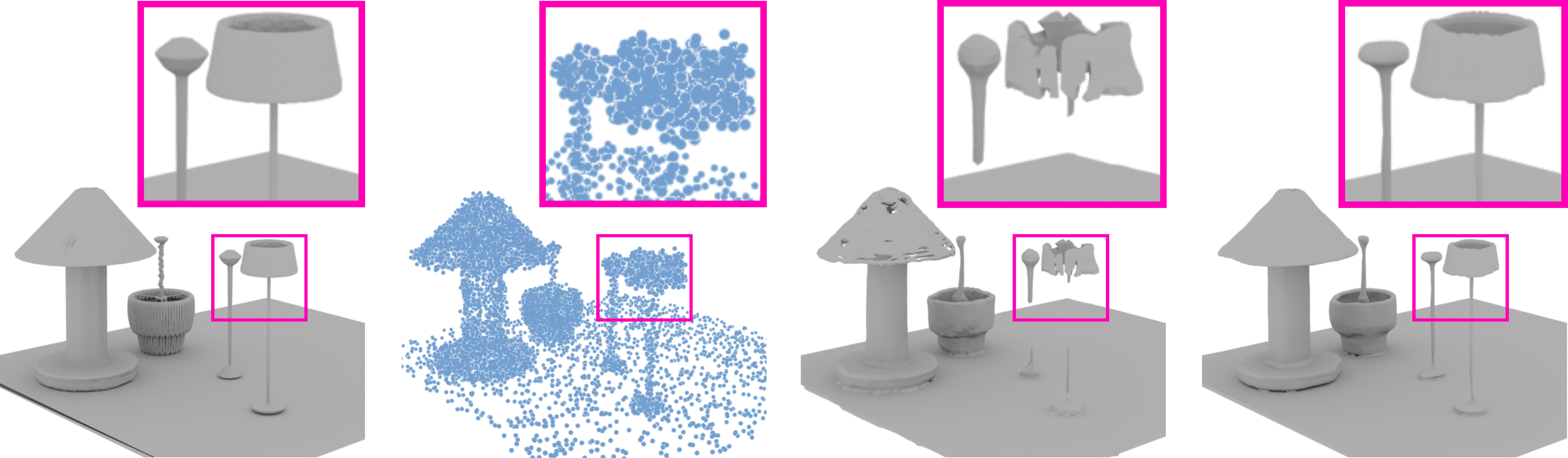}\vspace{-2mm}
    \hfill\subcaptionbox{Ground Truth}[0.25\linewidth]
    \hfill\subcaptionbox{Input (10,000 points)}[0.28\linewidth]
    \hfill\subcaptionbox{POCO~\cite{boulch2022poco} (36.8 sec)}[0.23\linewidth]
    \hfill\subcaptionbox{\textbf{ALTO (10.5 sec)}}[0.28\linewidth]
\hfill \vspace{-1mm}
    \hfill\caption{\textbf{Rethinking latent topologies for fast and detailed implicit 3D reconstructions}. Recent work (POCO CVPR'22~\cite{boulch2022poco}) has used latent encodings for each point to preserve 3D detail. We introduce ALTO, which can alternate between latent topologies like grid latents and point latents to speed up inference and recover more detail, like the 3D reconstruction of a thin lamp-post. Scene from~\cite{peng2020convolutional}.}
    \label{fig:teaser}
    \hfill \vspace{0mm}
    %\captionof{figure}{Test caption}
}]
{
\renewcommand{\thefootnote}{\fnsymbol{footnote}}
\footnotetext[1]{Equal contribution.}
}
% 
% \twocolumn[{
% \maketitle
% \centering
%  \includegraphics[width=\textwidth]{figures/F_teaser.pdf}
% \captionof{figure}{\textbf{Surface reconstruction of an input point cloud} from a ShapeNet Object-level Scene (left) and Scannet Large-scale Scene (right), along with runtime comparisons. Point Convolution for Surface Reconstruction (POCO) is a state-of-the-art comparison method~\cite{boulch2022poco}.}
% \label{fig:teaser}
% }]

\begin{abstract}
This work introduces alternating latent topologies (ALTO) for high-fidelity reconstruction of implicit 3D surfaces from noisy point clouds. Previous work identifies that the spatial arrangement of latent encodings is important to recover detail. One school of thought is to encode a latent vector for each point (point latents). Another school of thought is to project point latents into a grid (grid latents) which could be a voxel grid or triplane grid. Each school of thought has tradeoffs. Grid latents are coarse and lose high-frequency detail. In contrast, point latents preserve detail. However, point latents are more difficult to decode into a surface, and quality and runtime suffer. In this paper, we propose ALTO to sequentially alternate between geometric representations, before converging to an easy-to-decode latent. We find that this preserves spatial expressiveness and makes decoding lightweight. We validate ALTO on implicit 3D recovery and observe not only a performance improvement over the state-of-the-art, but a  runtime improvement of 3-10$\times$. Project website at \url{https://visual.ee.ucla.edu/alto.htm/}.
\end{abstract}
\section{Introduction}
\label{sec:intro}

% our method is general. can be applied to existing grid or point-based methods. 
% can be simply plugged in for other neural fields
% tri-plane-based: efficient but lacks communication between the UNets. 

Reconstructing surfaces from noisy point clouds is an active problem in 3D computer vision. Today, conditional neural fields offer a promising way to learn surfaces from noisy point clouds. Alternatives like voxel regression or mesh estimation are limited by cubic complexity and the requirement of a mesh template, respectively. Recent work has successfully used conditional neural fields to reconstruct 3D surfaces as an occupancy function. A conditional neural field takes as input a query coordinate and conditions this on a latent representation, e.g., feature grids. % of a noisy point cloud. 
The spatial expressiveness of the latent representation impacts the overall surface reconstruction quality. 

\begin{figure*}[t]
  \centering
%   \fbox{\rule{0pt}{3in} \rule{0.9\linewidth}{0pt}}
   \includegraphics[width=\linewidth]{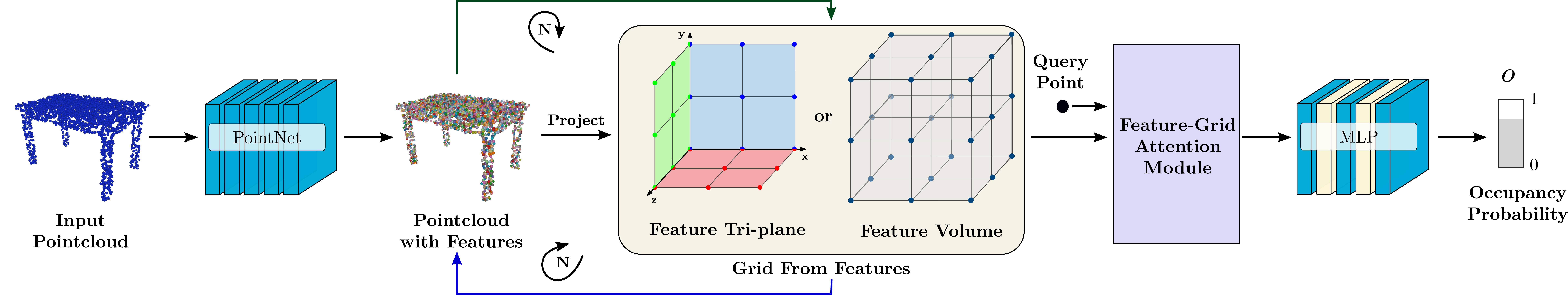}

   \caption{\textbf{An overview of our method.} Given input surface points, we obtain an implicit occupancy field with iterative alternation between features in the forms of points and 2D or 3D grids (\cref{sec:alternating}). Then we decode the occupancy values for query points with a learned attention-based interpolation from neighboring grids (\cref{sec:attention}).}
   \label{fig:pipeline}
\end{figure*}

To achieve spatial expression, a neural field is conditioned on a latent space of features (\textbf{latents}) from the conditional input. In 3D surface reconstruction the input point cloud is transformed into latents arranged in some topological structure. \textbf{Point latents} occur when each point in the input point cloud is assigned a latent vector~\cite{boulch2022poco}. \textbf{Triplane latents} are formed when point latents are projected into a 3-axis grid~\cite{lionar2021dynamic, peng2020convolutional}. The triplane latent is not as spatially expressive as freeform points, but the lower spatial complexity makes it easier to decode. \textbf{Voxel latents} are another type of grid latent where latents are arranged in a feature volume~\cite{peng2020convolutional, tang2021sa}. 
% A voxel latent is considered in-between the point and triplane in its spatial expressivity and ease of decoding. \achuta{zhen can you address gordon's comment and add 1-2 refs here?}

To reconstruct detailed surfaces, recent state-of-the-art methods try to preserve point latents as long as possible. Because point latents are spatially expressive, methods based on point latents are considered state-of-the-art for detailed surface reconstruction~\cite{boulch2022poco,erler2020points2surf}. However, using point latents in this way has some tradeoffs. It is difficult to correlate a query with the unstructured topology of a point-based latent space, placing a burden on the decoder. Results from POCO~\cite{boulch2022poco} are shown in~\cref{fig:teaser} where runtime and high-quality detail like thin lampposts remain out of reach.

In this paper, we seek to blend the upside of different latent topologies, while minimizing downside. We present an alternating latent topology (ALTO) method. In contrast to previous work, our method does not stay with either point~\cite{boulch2022poco} or grid latents~\cite{peng2020convolutional}, but instead alternates back and forth between point and grid latents before converging to a final grid for ease-of-decoding. 

% ALTO can convert grid features back into point features after convolutional operations on grids, so point features with aggregated neighborhood information are learned at finer granularity with just a simple per-point multi-layer perceptron (MLP). For 2D grids, the MLP also provides an extra benefit of promoting communication among three planar features. We then project point features back to grid features for convolution. The alternation makes these two forms of features complementary: fine-grained per-point feature learning lead to better grid features and convolutional grid features can efficiently propagate both global and local features for each point. 
Our method is general. We can plug-in the ALTO component to existing grid-based conditional models \cite{peng2020convolutional,Chibane2020CVPR} to boost detail recovery. While we have shown that our method can generate occupancy fields, we expect gain of high-fidelity details for other neural fields, such as semantic or affordance fields \cite{vora2021nesf, jiang2021synergies}, where similar conditional techniques can be adopted.

We summarize our \textbf{contributions} as follows:
\begin{itemize}\itemsep=0cm
    \item{We introduce an iterative technique to blend the strengths of different latent topologies for high-fidelity conditional neural fields generation.}
    \item{We propose an attention-based decoder that replaces naive linear interpolation of feature-grids or computationally expensive point-wise attention while keeping compute burden in check.}

    % \item {We improve runtime over existing occupancy decoders via an attention-based mechanism that replaces linear interpolation of feature-grids and computationally expensive point-wise attentions.}
    
            \item {We demonstrate performance and runtime improvements over the highest-quality previous method~\cite{boulch2022poco}, as well as performance improvements over all other baselines.}
    % \item {Performance and runtime improvements over the highest-quality previous method~\cite{boulch2022poco}. Performance improvements over all other baselines.}
\end{itemize}

% \begin{figure}[t]
%   \centering
%   \fbox{\rule{0pt}{3.5in} \rule{\linewidth}{0pt}}
%   %\includegraphics[width=0.8\linewidth]{egfigure.eps}

%   \caption{Maybe a teaser figure: our method is both fast and accurate compared with e.g. POCO and/or SA-ConvONet.}
% \end{figure}

    % \item We propose a simple point-grid alternating block that converts point cloud features iteratively between two previous commonly \emph{yet} separately used forms: per-point form and grid (either 2D or 3D) form;
    
    % \item We incorporate this block into a hourglass framework and the resulting point-grid alternating U-Net architecture enables more efficient alternation between the two form; and 
    % \item Extensive empirical results on both object and scene-level synthetic and real-world datasets demonstrate that our method can learn better implicit representation for surface reconstruction.
    % \item (TBD) A new dataset which extract occupancy labels from 3D front dataset.
    % \item (TBD) A new occupancy decoder using attention mechanism based on our learned feature forms.

\section{Related Work}
\label{sec:related}

%Representations for 3D scenes fall under three broad categories: explicit data structures, neural fields and conditional neural fields. Our work is most closely related to conditional neural fields, Section~\ref{ss:conditional}.
In this section, we discuss the most relevant literature on learning-based 3D reconstruction methods.
Based on their output, existing learning-based approaches can be categorized
as implicit or explicit-based representations. In this work, we primarily focus on implicit-based
representations as they are closely related to our method.

%\subsection{Explicit Data Structures}
\subsection{Explicit Representations}
\label{ss:explicit}

A common     shape representation is 2.5D depth maps, which can be inferred using 2D CNNs \cite{Kar2017NIPS, Hartmann2017ICCV, Paschalidou2018CVPR, Huang2018CVPR, Yao2018ECCV, Donne2019CVPR, Yao2019CVPR, Yu2020CVPR}.
However, 2.5D depth maps cannot capture the full 3D geometry. In contrast, voxels \cite{Maturana2015IROS, Wu2015NIPS, Brock2016ARXIV, Choy2016ECCV, Wu2016NIPS, Gadelha2017THREEDV, Rezende2016NIPS, Stutz2018CVPR, Xie2019ICCV, dai2020sg} naturally capture 3D object geometry,
by discretizing the shape into a regular grid. As voxel-based methods exhibit cubic space complexity that results
in high memory and computation requirements, several works tried to circumvent this with efficient space partitioning techniques \cite{Maturana2015IROS, Riegler2017CVPR, Tatarchenko2017ICCV, Haene2017ARXIV, Riegler2017THREEDV}.
Although these methods allow for increasing the voxel resolution and hence capturing more complex geometries,
their application is still limited. Recently, a promising new direction explored learning grid deformations
to better capture geometric details~\cite{Gao2020NIPS}. An alternative representation relies on pointclouds.
Point-based approaches \cite{Fan2017CVPR, Qi2017NIPS, Achlioptas2018ICML, Jiang2018ECCV, Thomas2019ICCV, Yang2019ICCV}
discretize the 3D space using points and are more light-weight and memory efficient. However, as they lack surface connectivity,
they require additional post-processing steps (\eg using Poisson Surface Reconstruction\cite{Kazhdan2006PSR})
for generating the final mesh. Instead, mesh-based methods~\cite{Liao2018CVPR, Groueix2018CVPR, Kanazawa2018ECCV, Wang2018ECCV, Yang2019ICCV, Gkioxari2019ICCV, Pan2019ICCV, Chen2019NIPS, Deng2020CVPR}
naturally yield smooth surfaces but they typically
require a template mesh~\cite{Wang2018ECCV}, which makes scaling them to arbitrarily complex
topologies difficult. Other works, proposed to also represent the geometry as an atlas of 
mappings~\cite{Groueix2018CVPR, Deng2020THREEDV, Ma2021CVPR}, which can result in
non-watertight meshes. To address limitations with learning explicit representations, 
implicit models~\cite{mescheder2019occupancy, Chen2019CVPR, Park2019CVPR,sitzmann2019siren } emerged as an alternative
more compact representation that yield 3D geometries at infinitely high resolutions using iso-surfacing operations
(\ie marching cubes). In this work, we capture 3D geometries implicitly,
using an occupancy field~\cite{mescheder2019occupancy}, as it faithfully can
capture complex topologies.

%Explicit data structures include point clouds, grids, voxels, and meshes. A \textbf{point cloud} organizes the geometric location of points into an array and are considered light-weight and efficient~\cite{qi2017pointnet,qi2017pointnet++}. In addition, point features can be processed at very fine granularity (i.e. at individual point level). However, point clouds cannot model topological relations. A \textbf{grid} is formed by projecting points into 2D or 3D planes, enabling conventional convolution kernels to learn local features~\cite{hua2018pointwise,xu2018spidercnn,atzmon2018point}. However, grids do not model the full volume, whereby one might appeal to a \textbf{voxel } data structure~\cite{choy20163d, maturana2015voxnet, dai2020sg, wu2016learning, wu20153d, xie2019pix2vox}. Unfortunately, voxel-based methods exhibit cubic space complexity, which limits performance. Alternatively, one can learn a \textbf{mesh}, consisting of vertices and faces~\cite{Liao2018CVPR,
%Groueix2018CVPR, Kanazawa2018ECCV, Wang2018ECCV, Yang2019ICCV, Pan2019ICCV}. Unfortunately, this typically requires a template mesh and arbitrary topologies are difficult to represent with a template mesh.  

\subsection{Neural Implicit Representations} 
\label{ss:implicit}

Unlike explicit representations that discretize the output space using voxels,
points or mesh vertices, implicit representations represent the 3D shape and appearance implicitly,
in the latent vectors of a neural network that learns a mapping between
a query point and a context vector to either a signed distance value
\cite{Park2019CVPR, Michalkiewicz2019ICCV, Atzmon2020CVPR, Gropp2020ICML,
Takikawa2021CVPR} or a binary occupancy value \cite{mescheder2019occupancy, Chen2019CVPR, Sitzmann2019NIPS}.
However, while these methods typically rely on a single global latent code, they cannot capture local details and
struggle scaling to more complicated geometries.
To address this, several works \cite{Saito2019ICCV, Xu2019NIPS, Saito2020CVPR} explored 
pixel-aligned implicit representations, that rely on both global and local image features
computed along a viewing direction. While, these approaches are able to capture fine-grained geometric details,
they rely on features that are computed from images, hence are limited to image-based inputs
with known camera poses.

Our work falls in the category of methods that perform 3D reconstructions from points. Among the first
to explore this direction were \cite{peng2020convolutional, Chibane2020CVPR,
Jiang2020CVPR}. To increase the expressivity of the underlying representation and to be able to capture
complex geometries, instead of conditioning on a global latent code, these works condition on local per-point
features. For example, Jiang \etal ~\cite{Jiang2020CVPR} leverage shape priors by conditioning on a 
patch-based representation of the point cloud. Other works \cite{peng2020convolutional,lionar2021dynamic,tang2021sa}, 
utilize grid-based convolutional features extracted from feature planes \cite{lionar2021dynamic},
feature volumes \cite{Chibane2020CVPR, tang2021sa} or both
\cite{peng2020convolutional}. An alternative line of work,
\cite{williams2022neural} introduce a test-time optimization mechanism to refine the per-point features predicted on a
feature volume. Concurrently, POCO~\cite{boulch2022poco}, propose to estimate per-point features, which are then
refined based on the per-point features of their neighboring points using an attention-based mechanism. A similar idea is also explored in Points2Surf~\cite{Erler2020ECCV} that introduces
a patch-based mechanism to decide the sign of the implicit function.  Our work is closely related to
POCO~\cite{boulch2022poco} that can faithfully capture higher-frequency details due its point-wise latent coding.
However, the lack of a grid-like structure places extra complexity on the attention-based aggregation module, 
which results in a higher computation cost. AIR-Net~\cite{giebenhain2021air} applied local attention to reduce the computation but limits its operating range to objects.
% and problems with generalizability, especially in areas with fine detail.
In this work, we propose conditioning on a hybrid representation of points and grid latents.
%It is reminescent of work from in 3D semantic segmentation that fuses points and voxels~\cite{liu2019point}.
In particular, instead of fusing points and grids, we demonstrate that it is the point and grid \emph{alternation}
between points and grids that enables recovery of more detail than POCO~\cite{boulch2022poco},
while reducing compute time by an order of magnitude.

\subsection{Obtaining Implicit Fields from Images} 
\label{ss:implicit}

As the previous methods require 3D supervision, several recent works
propose combining implicit representations with surface 
\cite{Niemeyer2020CVPR, Yariv2020NIPS} or volumetric \cite{Mildenhall2020ECCV}
rendering techniques in order to learn 3D object geometry and texture from
images. Among the most extensively used implicit representations are Neural
Radiance Fields (NeRF) \cite{Mildenhall2020ECCV} that combine an implicit representation
with volumetric rendering to perform a novel view synthesis. In particular, they employ
a neural network that maps a 3D point along a viewing direction to a color and a density value.
NeRFs are trained using only posed images. On the other hand \cite{Niemeyer2020CVPR, Yariv2020ARXIV}
combined occupancy fields~\cite{mescheder2019occupancy} and signed distance functions (SDFs) with
surface rendering in order to recover the geometry and appearance of 3D objects. 

Though we only demonstrate a case study in occupancy field in this paper, our proposed method is general and can be readily plugged into all kinds of other neural fields for better 3D point feature encoding. We leave that as future exploration.

\begin{figure*}[t]
%   \centering
%   \fbox{\rule{0pt}{3in} \rule{0.9\linewidth}{0pt}}
  \includegraphics[width=\linewidth]{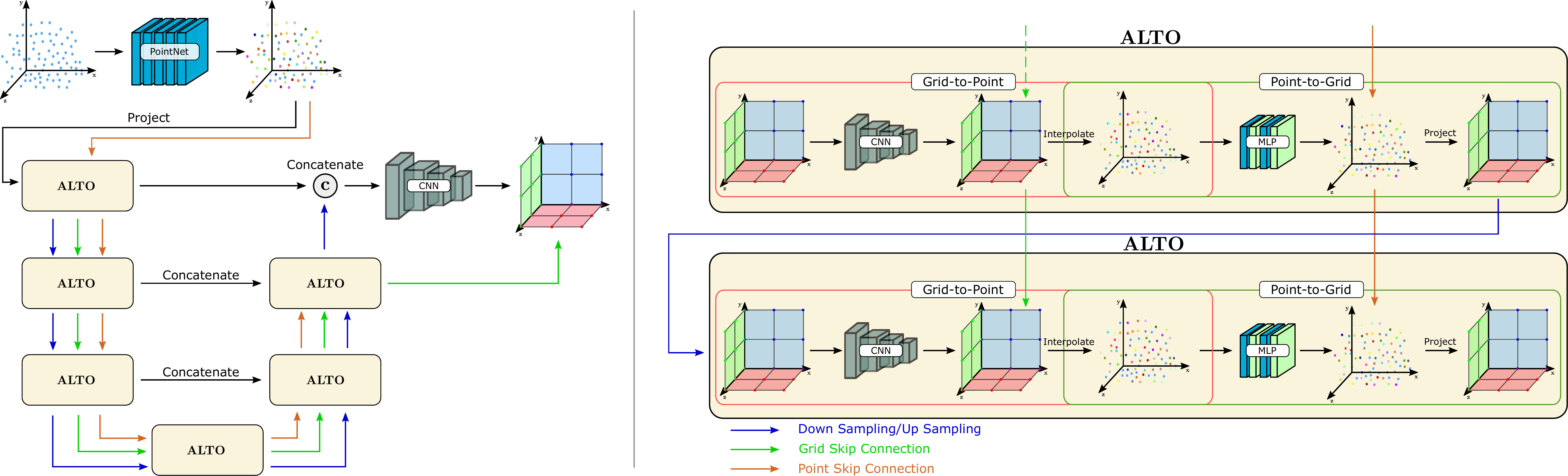} 
  \hfill\subcaptionbox{}[0.45\linewidth]
  \hfill\subcaptionbox{\label{fig:alto_consecutive}}[0.48\linewidth]
  
  \vspace{-.2cm}\caption{\textbf{An illustration of our ALTO encoder.} (a) As an example, we show the ALTO block instantiated by alternating between two latent topologies: point and triplanes via an ``in-network'' fashion, i.e. within each level of an hourglass framework U-Net.  `Concatenate' refers to concatenation of the ALTO block output triplane in the downsampling stage and the ALTO block input triplane in the corresponding upsampling stage. (b) We expand on ALTO block to illustrate the sequential grid-to-point and point-to-grid conversion. There are skip connections for both point and grid features between two consecutive levels in the ALTO U-Net. }
  \label{fig:alternation}
\end{figure*}
%  \zw{still need to iterate with Despi on a few changes of this fig.} \achuta{Rather than using Point-Grid we may want to do ALTO(Point, TriPlane) so it's written in a functional form?}.

%\begin{figure*}[t]
%  \centering
%   \fbox{\rule{0pt}{3in} \rule{0.9\linewidth}{0pt}}
%   \includegraphics[width=\linewidth]{figures/unet.JPG}

%   \caption{An overarching illustration of the general idea: point-grid alternating U-Net}
%   \label{fig:alternation}
%\end{figure*}
%\begin{figure*}[t]
%  \centering
%   \fbox{\rule{0pt}{3in} \rule{0.9\linewidth}{0pt}}
%   \includegraphics[width=\linewidth]{figures/unet_grid/unet_grid_2}
%
%   \caption{An overarching illustration of the general idea: point-grid alternating U-Net}
%   \label{fig:alternation}
%\end{figure*}

%Learning Continuous Environment Fields via Implicit Functions
\section{Method}

\label{sec:method}
There are three insights that motivate our approach: (1) conditioning on the right topology of the latent space is important; (2) previous neural fields for surface reconstruction condition on point or grid latents; (3) both point and grid latents have complementary strengths and weaknesses. Point latents are more spatially expressive but grid latents are easier to decode into a surface.

It might seem like a simple concatenation of point and grid latents would be sufficient. The problem is that point latents remain difficult to decode (even if concatenated with a grid latent). Therefore, our insight is to alternate between grid and point latents, and converge to a grid latent. For feature triplane latents, the alternation also permits communication between the individual planes, which in previous, grid-based works would have been fed into independent hourglass U-Nets~\cite{peng2020convolutional}. 

In this section, we introduce a version of ALTO as a point-grid alternating U-Net. An overview of the method is shown in~\cref{fig:pipeline}. We demonstrate how the convolutional grid form is learned in~\cref{sec:feature_grid}, our point-grid alternating network in~\cref{sec:alternating}, our attention-based decoder in~\cref{sec:attention} and training and inference in more detail in~\cref{sec:training}. 
 
\subsection{Convolutional Feature Grids}
\label{sec:feature_grid}

The input to our method is a noisy un-oriented point cloud $\mathcal{P}=\left\{\boldsymbol{p}_{i} \in \mathbb{R}^{3}\right\}_{i=1}^{S}$, where $S$ is the number of input points. We first use a shallow Point-Net~\cite{qi2017pointnet} to obtain the initial point features as in ConvONet~\cite{peng2020convolutional}. 
 These point features are then projected into three 2D grids (feature triplanes) $\{\mathbf{\Pi}_{xy}, \mathbf{\Pi}_{xz}, \mathbf{\Pi}_{yz}\} \in  \mathbb{R}^{H \times W \times d}$ or 3D grids (feature volumes) $\mathbf{V} \in  \mathbb{R}^{H \times W \times D \times d}$ before feeding into a 2D or 3D convolutional hourglass (U-Net) networks~\cite{cciccek20163d, ronneberger2015u}. $d$ is the number of feature channels. For feature volumes, we set $H = W = D = 64$ due to memory overhead of 3D-CNN and for feature triplanes, $H$ and $W$ can be set as high as 128 depending on the task.
 
% Seen from the lens of ConvONet~\cite{peng2020convolutional}, a key distinction is that we do not proceed with the grid latent, but instead iteratively alternate between grid and point latents to learn a fine-grained implicit representation for decoding occupancy.

\subsection{ALTO Latent to Blend Grid and Point Latents}
\label{sec:alternating}
Without loss of generality, we demonstrate ALTO in the context of blending grid and point latents. Note that 
naive concatenation of latents would not work, as the point latents are difficult to decode. The goal is to use ALTO to blend point latent characteristics into a grid latent via alternation. The alternating block is illustrated in~\cref{fig:alternation} and incorporated into a U-Net architecture. At each alternation, we first do grid-to-point conversion where convolutional grid features are transformed into point features, followed by point-to-grid conversion where extracted point features are transformed back into grid features for next alternation.

\textbf{Grid-to-Point Conversion:} At each alternation, to aggregate local neighborhood information, we use convolutional operations for the grid features. We then project each point $p$ orthographically onto the canonical planes and query the feature values through bilinear interpolation for 2D grid and trilinear interpolation for 3D grid. For triplane latents, we sum together the interpolated features from each individual plane.

\textbf{Point-to-Grid Conversion:} At each alternation, given the interpolated point features, we then process point features with an MLP in order to model individual point feature with finer granularity. For feature triplanes grid form, an MLP also gives an additional benefit of having individual plane features communicate with each other. The MLP is implemented with two linear
layers and one ReLU nonlinearity.
Projected point features falling within the same pixel or voxel cell will be aggregated using average pooling. If using triplane latents with each plane discretized at $H \times W$, this results in planar features with dimensionality $H \times W \times d$ 
or if using voxel latents we obtain dimensionality $H \times W \times D \times d$, where $d$ is the number of features.

We also adopt skip connections for both point features and grid features between two consecutive ALTO blocks, as illustrated in ~\cref{fig:alto_consecutive}.
The alternation needs to be implemented carefully to minimize runtime. Naively, we can alternate between triplanes or voxel latents using a U-Net and point latents using MLP, but that would require multiple network passes. Instead, we incorporate the point-grid alternating \emph{inside} each block of a U-Net, i.e. replacing original convolution-only block with ALTO block. We call this single U-Net, the ALTO U-Net. This also enables point and grid features blended at multiple scales and the number of alternation blocks depends on the number of levels of U-Net.

\subsection{Decoding ALTO latents using Attention}
\label{sec:attention}

As discussed in the previous section, ALTO provides a way to get a single latent that blends characteristics of different topologies. The advantage is that the final latent that ALTO converges to (hereafter, ALTO latent) can take on the topology that is easier to decode. 

For example, in the case of using ALTO to blend point and grid characteristics, we would like ALTO to converge to a final output in the simpler grid topology. Then, given the ALTO latent in grid form and any query point $\mathbf{q} \in \mathbb{R}^3$ in 3D space, our goal is to decode the learned feature and estimate the occupancy probability of each query point. ALTO benefits from attention on the decoder side. The ALTO latent is in grid form, but has spatial expressivity coming from the blended-in point latents. Standard grid latent decoding, e.g., bi-/tri-linear interpolation used in previous work~\cite{peng2020convolutional, lionar2021dynamic} would not preserve this spatial expressivity. 

\begin{figure}[t]
  \centering
%   \fbox{\rule{0pt}{3in} \rule{0.9\linewidth}{0pt}}
   \includegraphics[width=\linewidth]{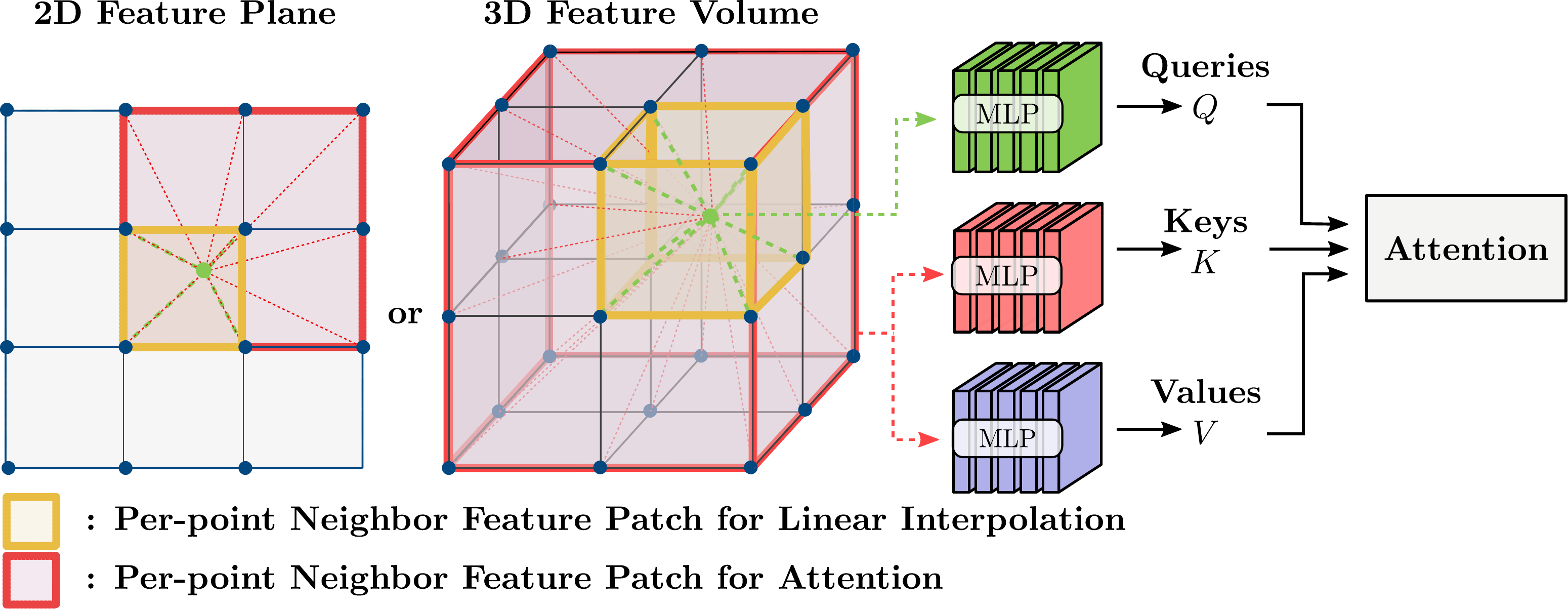}
   \caption{\textbf{Attention-based decoder on neighboring grids (2D or 3D).} To obtain features of each query point for decoding occupancy value, we use learned interpolation from neighboring grids that improves occupancy prediction, while being more efficient than expansive point-wise attention mechanism (e.g. POCO~\cite{boulch2022poco}).}
   \label{fig:attention}
\end{figure}

To decode an ALTO latent, we propose an efficient attention-based mechanism
to replace the previous approach of linear interpolation on feature grids. While attention is not new, we leverage \emph{grid latent attention} to avoid heavy runtime issues of \emph{point latent attention}~\cite{boulch2022poco} that applies attention over a point-wise 3D neighborhood. As illustrated in \cref{fig:attention}, 
% besides the traditional linear interpolation, 
we consider the nearest grids (indices denoted as $\mathcal{N}(\mathbf{q})$), where $|\mathcal{N}|=9$ for triplane representation and $|\mathcal{N}|=27$ for volume representation, we call these areas as per-point neighbor feature patches $\mathbf{C}_{\{i \in \mathcal{N}\}}$. We define the query $Q$, key $K$, and value $V$ for our attention as follows:
\begin{equation}
\begin{aligned}
Q &= \text{MLP}(\mathbf{\psi(q)}),\\
K &= \text{MLP}(\mathbf{C}_{\{i \in \mathcal{N}(\mathbf{q})
\}}),\\
V &= \text{MLP}(\mathbf{C}_{\{i \in \mathcal{N}(\mathbf{q})
\}}),
\end{aligned}
\end{equation}
where $\mathbf{\psi(q)}$ is the linear interpolated feature value of the query points. Additionally, we  compute the displacement vector $\mathbf{d} \in \mathbb{R}^2$ or $\mathbb{R}^3$ which represents the spatial relationship between the projected query point coordinate and the nearest feature grid points. We use the subtraction relation~\cite{zhao2021point} in our attention scoring function:

\begin{equation}
A = \text{softmax}(\text{MLP}((Q-K)+\gamma(\mathbf{d}))),
\end{equation}
where $\gamma(\mathbf{d})$ works as a learnable positional encoding. In our implementation, $\gamma$ is an MLP with two fully-connected layers and activated by ReLU. We compute the attention-based interpolated per-point feature $F$ as:
\begin{equation}
F = A \odot (V + \gamma(\mathbf{d})).
\end{equation}
Note that the same positional encoding from above is added to $V$ and $\odot$ denotes the element-wise product operation. For the case of triplane representation, we use a single-head attention to extract the feature $F$ from each individual plane. The per-triplane features are then concatenated and used for the occupancy prediction.
For the case of the volume representation, we use multi-head attention for $h$ independently learned subspaces of $Q$, $K$, and $V$, where $h$ is a hyperparameter varying based on the experiment. Additional details are provided in the supplementary.
\begin{table*}[t]
    % \vskip 0.05in
  \begin{center}
  \resizebox{\linewidth}{!}{
  \begin{tabular}{lcccccccccccc} 
    \toprule
       \multirow{1}{*}{} & \multicolumn{4}{c}{\centering input points 3K} & \multicolumn{4}{c}{\centering input points 1K} & \multicolumn{4}{c}{\centering input points 300} \\
    \cmidrule(l){2-5} \cmidrule(l){6-9} \cmidrule(l){10-13} 
    Method & IoU $\uparrow$ & Chamfer-$L_1 \downarrow$ & NC$\uparrow$ & F-score$\uparrow$ & IoU $\uparrow$ & Chamfer-$L_1 \downarrow$ & NC$\uparrow$ & F-score$\uparrow$ & IoU $\uparrow$ & Chamfer-$L_1 \downarrow$ & NC$\uparrow$ & F-score$\uparrow$  \\
    \midrule
    ONet~\cite{mescheder2019occupancy} & 0.761 & 0.87 & 0.891 & 0.785 & 0.772 & 0.81 & 0.894 &  0.801 & 0.778 & 0.80 & 0.895 &0.806 \\
%   SPSR~\cite{kazhdan2013screened} &  \\
%   SPSR trimmed~\cite{kazhdan2013screened} &  \\
  ConvONet~\cite{peng2020convolutional} &0.884 & 0.44 & 0.938 & 0.942 & 0.859 & 0.50 & 0.929 & 0.918 &0.821 &0.59 & 0.907 & 0.883\\
%   DP-ConvONet~\cite{lionar2021dynamic} & 0.895 & 0.42 & 0.941 & 0.952 & \\
  POCO~\cite{boulch2022poco} & 0.926 & \textbf{0.30} & 0.950 & \textbf{0.984} & 0.884 & 0.40 & 0.928 & 0.950 & 0.808 & 0.61 & 0.892 & 0.869 \\
  \midrule
  ALTO (Encoder Only) &  \textbf{0.931} & \textbf{0.30} & 0.950 & 0.981 & 0.889 & 0.39 & 0.932 & 0.951 & 0.842 & 0.52 & 0.908 & 0.903\\
  ALTO &  0.930 & \textbf{0.30} & \textbf{0.952} & 0.980 & \textbf{0.905} & \textbf{0.35} & \textbf{0.940} &  \textbf{0.964} & \textbf{0.863} & \textbf{0.47} & \textbf{0.922} & \textbf{0.924} \\
    \bottomrule
  \end{tabular}}
 \end{center}\vspace{-5mm}
  \caption{\textbf{Performance on ShapeNet with various point density levels.} Input noisy point cloud with 3K, 1K and 300 input points from left to right. ALTO is our proposed method and ALTO (Encoder only) is an ablation that uses only our encoder with a non-attention based decoder.}
  \label{tab:shapenet}
\end{table*}

\begin{figure}
     \includegraphics[width=\linewidth]{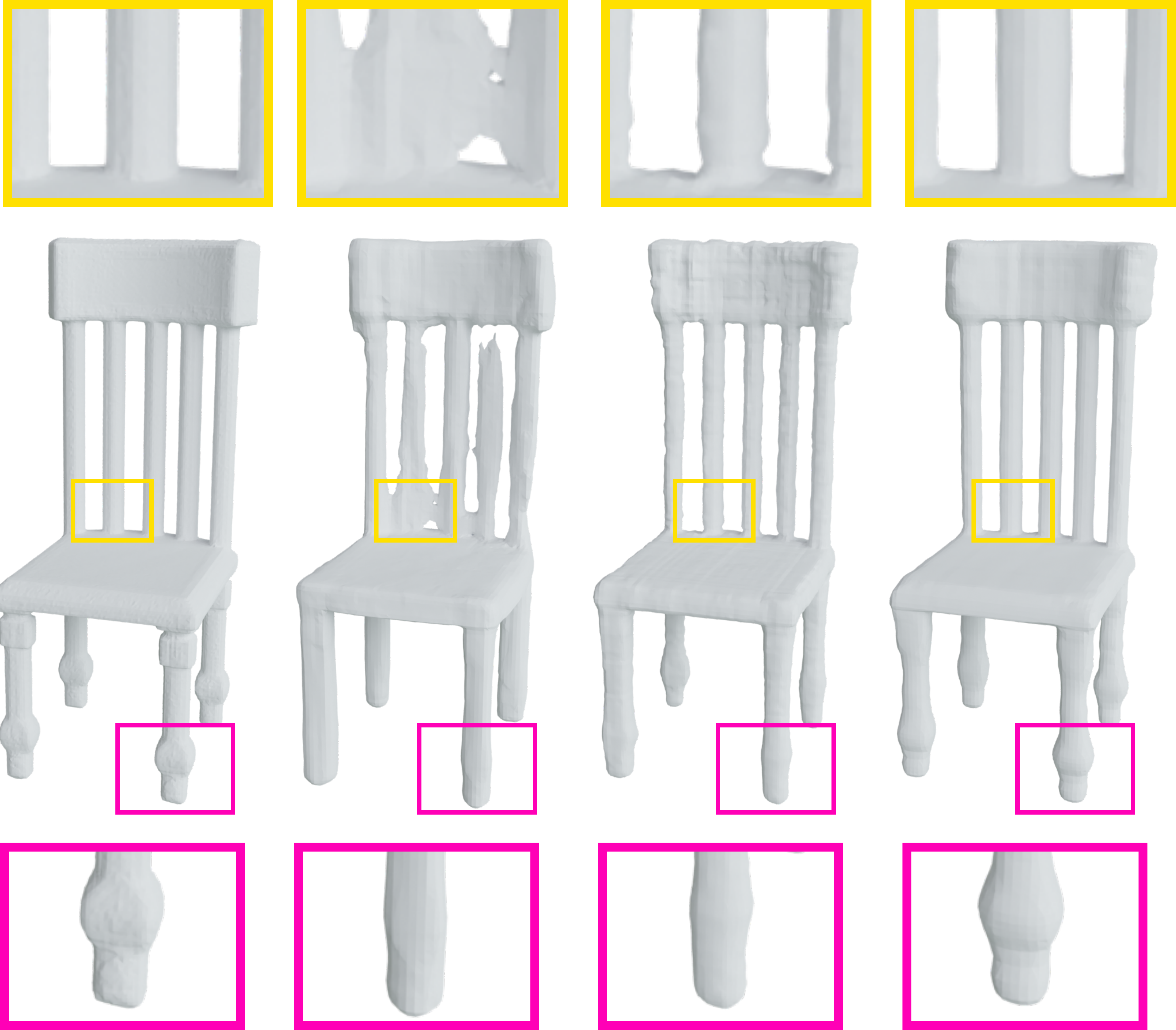}
      \hfill\subcaptionbox{GT}[0.22\linewidth]
    \hfill\subcaptionbox{Grid Latent~\cite{peng2020convolutional}}[0.28\linewidth]
    \hfill\subcaptionbox{~\cref{sec:alternating}\\\centering ``ALTO (Enc.)''}[0.24\linewidth]
    \hfill\subcaptionbox{~\cref{sec:alternating}+\ref{sec:attention}\\\centering ``ALTO''\label{fig:ablationd}}[0.27\linewidth]
    
    \caption{\textbf{Ablation analysis on ShapeNet.} Note the top inset showing the poles in the chair back (yellow). ALTO (Enc.) is ALTO (Encoder Only) and uses the latent space encoding proposed in~\cref{sec:alternating} with a standard decoder. The full ALTO method includes also the attention-based decoder in~\cref{sec:attention}.}
    \label{fig:ablation}
\end{figure}
Finally, we predict the occupancy of $\mathbf{q}$ using a small fully-connected occupancy network:
\begin{equation}
f_{\theta}(F) \rightarrow [0,1].
\end{equation}
The network $f_{\theta}$ consists of several ResNet blocks as in~\cite{peng2020convolutional}. The major difference to the original occupancy decoder in~\cite{peng2020convolutional} is that we do not bring in the absolute 3D coordinate of $\mathbf{q}$ as input since it theoretically breaks the translational equivalence property.

This concludes our description of our latent space encoding and attention decoding. In~\cref{fig:ablation} observe that the architectures we have proposed progressively improve detail from a standard grid formulation.

\subsection{Training and Inference}
\label{sec:training}

At training time, we uniformly sample query points $\mathcal{Q}$ and minimize the binary cross-entropy loss between the predicted occupancy value and ground-truth occupancy values written as: 
\begin{equation}
\mathcal{L}\left(\hat{o}_{\mathbf{q}}, o_{\mathbf{q}}\right) \! = \! \!- \!\!\sum_{\mathbf{q}\in \mathcal{Q}}\left[o_{\mathbf{q}} \log \left(\hat{o}_{\mathbf{q}}\right)\!+\!\left(1\!-\!o_{\mathbf{q}}\right) \log \left(1\!-\!\hat{o}_{\mathbf{q}}\right)\right]
\end{equation}
Our model is implemented in PyTorch~\cite{paszke2019pytorch} and uses the Adam optimizer~\cite{kingma2014adam} with a learning rate of $10^{-4}$. During inference, we use a form of Marching Cubes~\cite{lorensen1987marching} to obtain the mesh.
\section{Experimental Evaluation}
\label{sec:results}

\subsection{Datasets, Metrics, and Baselines}
\textbf{Object Level Datasets:} For evaluation on object-level reconstruction, we use ShapeNet~\cite{chang2015shapenet}.
% , ABC~\cite{koch2019abc}, Famous\cite{erler2020points2surf} and Things10k~\cite{erler2020points2surf}.
In particular, ShapeNet~\cite{chang2015shapenet} contains watertight meshes of object shapes in 13 classes. For fair comparison, we use the same train/val splits and 8500 objects for testing as described in~\cite{boulch2022poco,peng2020convolutional}. Points are obtained by randomly sampling 
% 3000 points 
from each mesh and adding Gaussian noise with zero mean and standard deviation of 0.05. 
% For ABC, Famous and Things10k, we follow a similar protocol, as discussed in the supplement.

\textbf{Scene-Level Datasets:} For scene level evaluation, we use Synthetic Rooms dataset~\cite{peng2020convolutional} and ScanNet-v2~\cite{Dai2017CVPR}. In total, we use 5000 synthetic room scenes with walls, floors and ShapeNet objects randomly placed together. We use an identical train/val/test split as prepared in prior work~\cite{peng2020convolutional,boulch2022poco}, and Gaussian noise with zero mean and 0.05 standard deviation. ScanNet-v2 contains 1513 scans from real-world scenes that cover a wide range of room types. Since the provided meshes in ScanNet-v2 are not watertight, models are trained on the Synthetic Rooms dataset and tested on ScanNet-v2, which also enables some assessment of Sim2Real performance of various methods.

\textbf{Evaluation Metrics:} The quantitative evaluation metrics used in data tables are standard metrics that enable us to form comparisons to prior work. These include: volumetric IoU, Chamfer-$L_{1}$ distance $\times 10^2$, and normal consistency (NC). A detailed definition of each metric can be found in the Occupancy Networks paper~\cite{mescheder2019occupancy}. We also include an F-Score metric~\cite{tatarchenko2019single} with threshold value $1\%$. 

\textbf{Baselines for Comparison:} As noted by Boulch et al.~\cite{boulch2022poco}, baseline methods often perform better in the settings of the original papers. In the same spirit, we thus strictly adapt the protocol of the state-of-the-art (SOTA) paper POCO~\cite{boulch2022poco} for evaluation protocol. In addition to POCO, other baselines we include are  SPSR~\cite{kazhdan2013screened}, ONet~\cite{mescheder2019occupancy} and ConvONet~\cite{peng2020convolutional}.
% , and Points2Surf~\cite{erler2020points2surf}. 
Note that we omit NFK~\cite{williams2022neural} from our evaluations, as they have not made their code publicly available.

\textbf{Our Method:} ``Our method'' is ALTO. ALTO combines~\cref{sec:alternating}~+~\cref{sec:attention}, and is shown in~\cref{fig:ablationd}. Figures/tables use ALTO to denote the proposed method. If we are considering an ablation analysis we will use ALTO with parentheses, e.g., ``ALTO (Encoder Only)'', and the table caption will specify the ablation. To demonstrate ALTO with different latent topologies , for object-level reconstruction, we use alternation between point and feature triplanes and the resolution of initial individual plane $H = W = 64$. For scene-level reconstruction, we use alternation between point and feature volumes and the resolution $H = W = D = 64$.

\begin{figure}[t]
%   \fbox{\rule{0pt}{3in} \rule{\linewidth}{0pt}}
 \includegraphics[width=\linewidth]{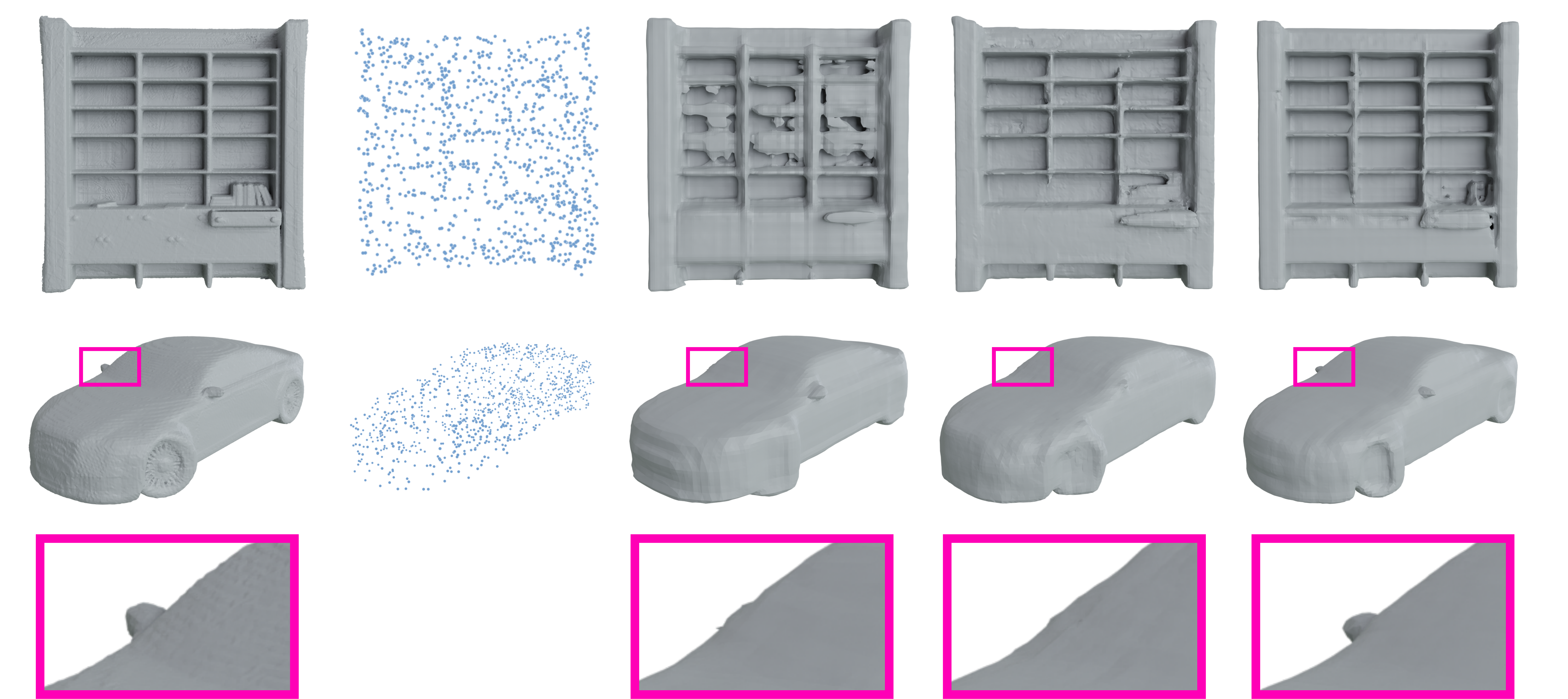}\vspace{-2mm}
       \hfill\subcaptionbox{GT}[0.20\linewidth]
    \hfill\subcaptionbox{Input (1K)}[0.21\linewidth]
    \hfill\subcaptionbox{CONet~\cite{peng2020convolutional}}[0.21\linewidth]
    \hfill\subcaptionbox{POCO~\cite{boulch2022poco}}[0.20\linewidth]
       \hfill\subcaptionbox{\textbf{ALTO}}[0.14\linewidth]
      
     \vspace{-.2cm}\caption{\textbf{Object-level comparisons on ShapeNet.} On the car, ALTO recovers the detail of having both side mirrors.}
  \label{fig:shapenet}
\end{figure}

\begin{figure*}[t]
%   \fbox{\rule{0pt}{6in} \rule{0.9\linewidth}{0pt}}
   \includegraphics[width=\linewidth]{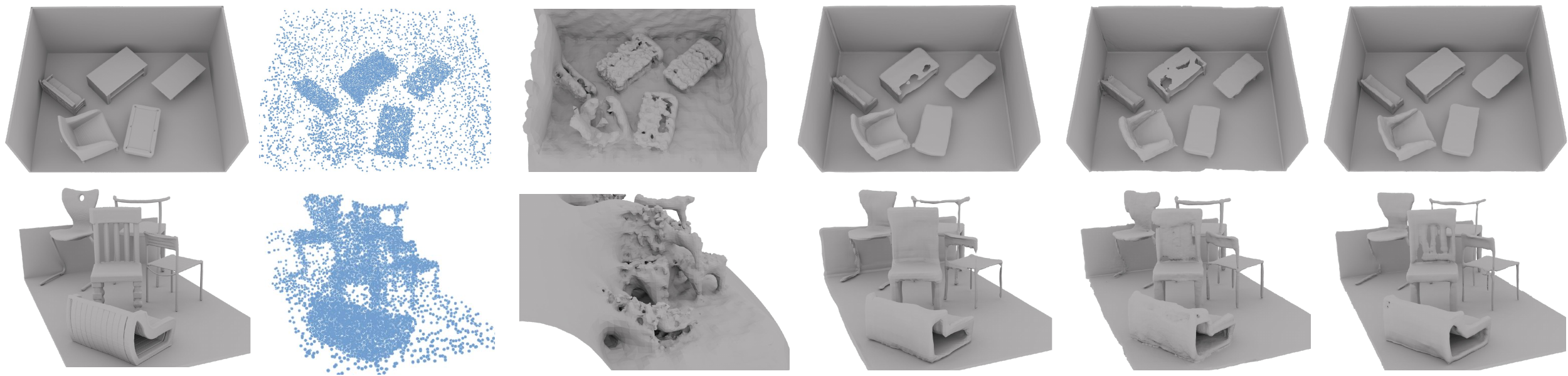}\vspace{-2mm}
\hfill\subcaptionbox{Ground Truth}[0.17\linewidth]
\hfill\subcaptionbox{Input Points}[0.14\linewidth]
\hfill\subcaptionbox{SPSR~\cite{kazhdan2013screened}}[0.22\linewidth]
\hfill\subcaptionbox{ConvONet~\cite{peng2020convolutional}}[0.15\linewidth]
\hfill\subcaptionbox{POCO~\cite{boulch2022poco}}[0.18\linewidth]
\hfill\subcaptionbox{\textbf{ALTO}}[0.13\linewidth]

\caption{\textbf{Qualitative comparison on scene-level reconstruction Synthetic Room Dataset.} Learning-based methods are trained and tested on 10K noisy points. ALTO can reconstruct the (top scene) double-deck table and (bottom scene) details in the chair.}
  \label{fig:syntheticroom}
%   \achuta{specify noise, e.g.g as pct of volume size noise - see previous paper.}
  % \achuta{(a) GT (b) Input .... (f) ours.}
\end{figure*}

\subsection{Results of Object-level Reconstruction}
\label{sec:object}

\textbf{Qualitative Object-level Comparisons:} Qualitative results of object-level reconstruction are provided in~\cref{fig:shapenet}. We observe that~\cite{peng2020convolutional} obtains a blurry reconstruction. Note that the width of the bookcase dividers are thicker and there are spurious blobs on the shelves. The SOTA baseline~\cite{boulch2022poco} is able to recover some detail, such as the wheel geometry in the car, but loses both side mirrors in the reconstruction. ALTO seems to have a higher fidelity reconstruction due to combining ideas from point and grid latents. 

\textbf{Quantative Object-level Comparisons:} ALTO's performance metrics at various point density levels are listed in~\cref{tab:shapenet}, for 3K, 1K and 300 input points. When point clouds are sparser, ALTO performs better than POCO on all four metrics. At high point density, ALTO outperforms POCO on three of four metrics. An ablation of just using the ALTO latent encoding and a traditional interpolating decoder~\cite{peng2020convolutional} is also conducted in the table. 

\begin{table}[t]

    % \vskip 0.05in
  \begin{center}
  \resizebox{\linewidth}{!}{
  \begin{tabular}{lccccccc} 
    \toprule
  
    Method & IoU $\uparrow$ & Chamfer-$L_1 \downarrow$ & NC$\uparrow$ & F-score$\uparrow$  \\
    \midrule
%   3D-CNN augmented & \textbf{3.32} & \textbf{4.01} &  \textbf{0.50} & \textbf{3.01}\\
%   3D-CNN baseline & 3.53 & 4.24 &0.43 & 2.75\\
   ONet~\cite{mescheder2019occupancy} & 0.475 &2.03& 0.783 & 0.541\\
  SPSR~\cite{kazhdan2013screened} & - & 2.23 & 0.866 & 0.810 \\
  SPSR trimmed~\cite{kazhdan2013screened} & - & 0.69 & 0.890 & 0.892 \\
  ConvONet~\cite{peng2020convolutional} & 0.849 & 0.42 & 0.915 & 0.964  \\
  DP-ConvONet~\cite{lionar2021dynamic} & 0.800 & 0.42 & 0.912 & 0.960  \\
  POCO~\cite{boulch2022poco} & 0.884 & 0.36 & 0.919 & 0.980
 \\
  \midrule
%   Ours ($ 3\times128^2$) &  0.834 & 0.43 & 0.906 & 0.960

% \\
%   Ours ($64^3$) &  0.903 & 0.36 & 0.920 & 0.978 \\
%%% 4heads hidden 128
ALTO & \textbf{0.914} & \textbf{0.35} & \textbf{0.921} & \textbf{0.981}\\
%%% 1head hidden 32 (1105 poco_mc)
%Ours (w/ our decoder, $64^3$) & \textbf{0.912} & \textbf{0.35} & \textbf{0.921} & \textbf{0.981}\\

    \bottomrule
  \end{tabular}}\vspace{-3mm}
 \end{center}
  \caption{\textbf{Synthetic Room Dataset.} Input points 10K with noise added. Boldface font represents the preferred results.}
   \label{tab:room}
\end{table}

\begin{table}[t]
    % \vskip 0.05in
  \begin{center}
  \resizebox{0.85\linewidth}{!}{
  \begin{tabular}{lccccccc} 
    \toprule
  
    Method & IoU $\uparrow$ & Chamfer-$L_1 \downarrow$ & NC$\uparrow$ & F-score$\uparrow$  \\
    \midrule
  ConvONet~\cite{peng2020convolutional} & 0.818 & 0.46 & 0.906 & 0.943 \\
  POCO~\cite{boulch2022poco} & 0.801 & 0.57 & 0.904 & 0.812  \\
  \midrule
  ALTO & \textbf{0.882} & \textbf{0.39} & \textbf{0.911} & \textbf{0.969} \\
    \bottomrule
  \end{tabular}}\vspace{-3mm}
 \end{center}
      \caption{\textbf{Performance on Synthetic Room Dataset (sparser input point cloud with 3K input points).} Boldface font represents the preferred results.}
\label{tab:room_sparse}
\end{table}

\subsection{Scene-level Reconstruction}
\label{sec:scene}

\textbf{Qualitative Scene-level Comparisons:} ALTO achieves detailed qualitative results compared to baselines. ~\cref{fig:syntheticroom} depicts scene-level reconstruction on the Synthetic Room dataset introduced in~\cite{peng2020convolutional}. In the first row of~\cref{fig:syntheticroom} the baselines of ConvONet and POCO both have holes in the coffee table. In the second row of~\cref{fig:syntheticroom} the high-frequency deatil in the wooden slats of the chair is fully blurred out by ConvONet. The advantages of ALTO are even more apparent for fine detail, such as the thin lamp-posts shown in ~\cref{fig:teaser}. ALTO reduces the quantization effect due to the grid discretization in the grid form by by using iterative alternation between grid and point latents, encoding more fine-grained local features for conditional occupancy field generation.

\textbf{Quantitative Scene-level Comparisons:} ALTO scores higher on quantitative values for scene-level metrics, shown in~\cref{tab:room} and~\cref{tab:room_sparse}. In the sparse setting, for the baselines methods, we find that ConvONet~\cite{peng2020convolutional} is quantitatively superior to the SOTA of POCO~\cite{boulch2022poco} because oversmoothing tends to improve quantitative results on noisy point clouds. Nonetheless, ALTO performs better than both baselines because ALTO limits spurious noise without resorting to as much oversmoothing.

\subsection{Real-world Scene Generalization}
\label{sec:real}

\begin{figure*}[t]
%   \fbox{\rule{0pt}{3in} \rule{0.9\linewidth}{0pt}}

  \includegraphics[width=\linewidth]{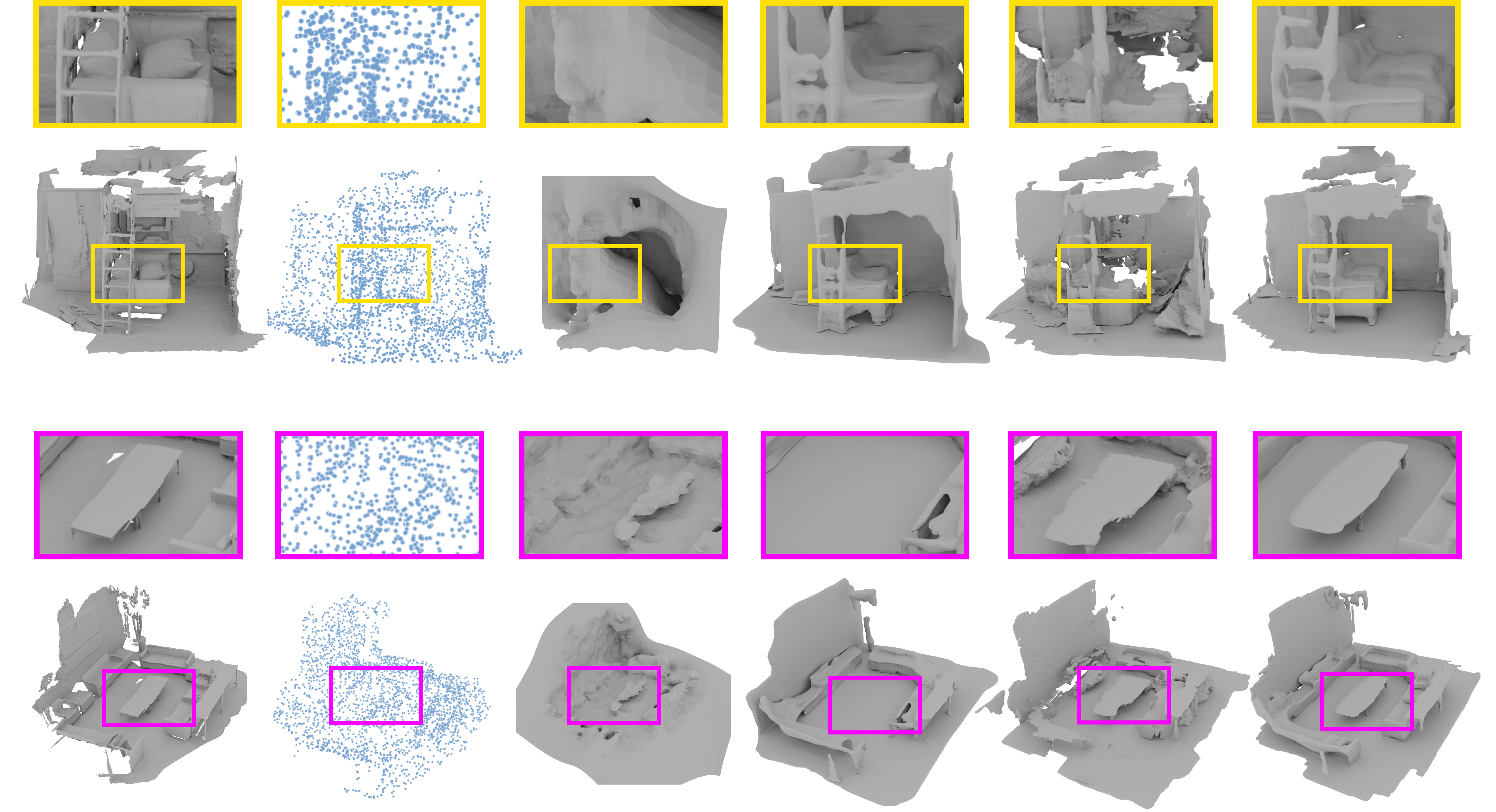}\vspace{-3mm}
\hfill\subcaptionbox{Ground Truth}[0.17\linewidth][l]
\hfill\subcaptionbox{Input Points}[0.16\linewidth]
\hfill\subcaptionbox{SPSR~\cite{kazhdan2013screened}}[0.18\linewidth][l]
\hfill\subcaptionbox{ConvONet~\cite{peng2020convolutional}}[0.14\linewidth]
\hfill\subcaptionbox{POCO~\cite{boulch2022poco}}[0.19\linewidth]
\hfill\subcaptionbox{\textbf{ALTO}}[0.13\linewidth]

  \caption{\textbf{Cross-dataset evaluation of ALTO and baselines by training on Synthetic Rooms~\cite{peng2020convolutional} and testing on real-world ScanNet-v2~\cite{Dai2017CVPR}}. Note the large conference-room table is missing in ConvONet~\cite{peng2020convolutional} (purple inset). The ladder (yellow inset) is a high-frequency surface and we believe our method is qualitatively closest. Please zoom in if browsing with PDF.} 
  \label{fig:scannet}
\end{figure*}

\begin{table}[t]
    % \vskip 0.05in
  \begin{center}
  \resizebox{\linewidth}{!}{
  \begin{tabular}{lccccccc} 
    \toprule
   \multirow{1}{*}{} & \multicolumn{2}{c}{\centering $N_{\text{Train}}$=10K, $N_{\text{Test}}$=3K} & \multicolumn{2}{c}{\centering $N_{\text{Train}}$=$N_{\text{Test}}$=3K} & \\
    \cmidrule(l){2-3} \cmidrule(l){4-5} 
    Method &  Chamfer-$L_1 \downarrow$  & F-score$\uparrow$& Chamfer-$L_1 \downarrow$  & F-score$\uparrow$ \\
    \midrule
%   ONet~\cite{mescheder2019occupancy} & 0.398 & 0.390\\
%   SPSR~\cite{kazhdan2013screened} & 0.293 & 0.731 \\
%   SPSR-trimmed~\cite{kazhdan2013screened} & 0.086 & 0.847 \\
  ConvONet~\cite{peng2020convolutional} & 1.01 & 0.719 &1.16 & 0.669\\
  POCO~\cite{boulch2022poco} & 0.93 & 0.737 & 1.15 & 0.667 \\ % 0.798\\
  \midrule
%   Ours & 0.95 & 0.745 & 0.98 & 0.827\\
  ALTO     & \textbf{0.87} & \textbf{0.746} & \textbf{0.92} & \textbf{0.726}\\
    \bottomrule
  \end{tabular}}
 \end{center}\vspace{-3mm}
  \caption{\textbf{ScanNet-v2.} We test the generalization capability of all the methods on real-world scans ScanNet using models trained on both the same input points of synthetic dataset as test set ($N_{\text{Train}}$=$N_{\text{Test}}$=3K) and different point density level ($N_{\text{Train}}$=10K, $N_{\text{Test}}$=3K). Boldface font represents the preferred results.   }
   \label{tab:scannet}
\end{table}

A final experiment in the main paper is to assess the performance of our model in real-world scans from ScanNet-v2~\cite{Dai2017CVPR}. All models are trained on Synthetic Rooms and tested on ScanNet-v2 to demonstrate generalization capability of our method along with baselines. We demonstrate the qualitative results of the setting where models trained
on both the same input points of synthetic dataset as ScanNet test set
($N_{\text{Train}}$=$N_{\text{Test}}$=3K) in~\cref{fig:scannet}. Our method is qualitatively superior to SPSR~\cite{kazhdan2013screened}, ConvONet~\cite{peng2020convolutional}, and POCO~\cite{boulch2022poco}. As in the Synthetic Rooms dataset, we observe that ConvONet oversmooths surfaces (sometimes causing entire objects to disappear, like the conference table in the purple inset of~\cref{fig:scannet}). In contrast, POCO retains some detail but is noisier. The quantitative results in~\cref{tab:scannet} are consistent with qualitative results. 
The cross point density test results ($N_{\text{Train}}$=10k, $N_{\text{Test}}$=3k) also demonstrates the superiority of our method on generalization when there are abundant input points in synthetic dataset used for training and low point density in real-world inference.

\begin{table}[t]
   
    % \vskip 0.05in
  \begin{center}
  \resizebox{0.8\linewidth}{!}{
  \begin{tabular}{lccccccc} 
    \toprule
  
    Method & $\#$ Parameters &  Inference time (s) \\
    \midrule
    % ConvONet ($3 \times 128^2$) & 7,812,705 & 1.80 & 0.97\\
    ConvONet~\cite{peng2020convolutional} & 4,166,657 & 1.6 \\ 
    POCO~\cite{boulch2022poco} & 12,790,454 & 36.1\\
    % \midrule
    % Ours ($3 \times 128^2$) & 2,713,889 & 1.74 & 4.86\\
    % Ours  ($64 ^3$) & 4,778,433 & 2.70 &4.48 \\
    % Ours (w/ our decoder, $3 \times 128^2$) & 2,763,073 & 3.00  & 6.93\\
    ALTO & 4,787,905&  3.6 \\
    \bottomrule
  \end{tabular}}
 \end{center}
  \caption{\textbf{Runtime comparison.} We report the number of parameters and inference time corresponding to~\cref{fig:scannet}. ALTO is much faster than POCO and recovers more detail~\cite{boulch2022poco}. ALTO is also only slightly slower than fast methods that are not as spatially expressive~\cite{peng2020convolutional}.}
   \label{tab:runtime}
\end{table}
% results on ScanNet-v2 and Matterport3D  

% \begin{table}[t]
%     \label{tab:ubfc}
%     % \vskip 0.05in
%   \begin{center}
%   %\resizebox{\linewidth}{!}{
%   \begin{tabular}{lccccccc} 
%     \toprule
  
%     Method &  Chamfer-$L_1 \downarrow$  & F-score$\uparrow$  \\
%     \midrule
%   ONet~\cite{mescheder2019occupancy} & 0.398 & 0.390\\
%   SPSR~\cite{kazhdan2013screened} & 0.293 & 0.731 \\
%   SPSR-trimmed~\cite{kazhdan2013screened} & 0.086 & 0.847 \\
%   ConvONet~\cite{peng2020convolutional} & 0.094\\
%   POCO~\cite{boulch2022poco} & \\
%   \midrule
%   Ours & \\
%     \bottomrule
%   \end{tabular}
%  \end{center}
%   \caption{\textbf{Performance on ScanNet-v2.} Boldface font represents the preferred results.}
% \end{table}

% \begin{figure}[t]
%   \centering
%   \fbox{\rule{0pt}{2in} \rule{0.9\linewidth}{0pt}}
%   %\includegraphics[width=0.8\linewidth]{egfigure.eps}

%   \caption{a plot showing various densities}
%   \label{fig:onecol}
% \end{figure}

\section{Discussion and Conclusion}

In summary, this paper has adopted a different philosophy from the SOTA in surface detail recovery. We rely neither on point latents~\cite{boulch2022poco} or grid latents~\cite{peng2020convolutional} alone, but alternate between topologies. The output of ALTO is a spatially expressive latent that is also topologically easy-to-decode into a 3D surface. This breaks a Pareto tradeoff that previous works have posited between spatial expressiveness and decoding complexity. For this reason, it is not surprising that our method reconstructs more detailed 3D surfaces with faster runtimes than state-of-the-art (\cref{fig:teaser} and~\cref{tab:runtime}).

% \begin{table}[t]
%     \label{tab:ubfc}
%     % \vskip 0.05in
%   \begin{center}
%   \resizebox{\linewidth}{!}{
%   \begin{tabular}{lccccccc} 
%     \toprule
  
%     Method & $\#$ Parameters & Memory (GB) &  Inference time (s) \\
%     \midrule
%     ConvONet ($3 \times 128^2$) & 7,812,705 & 1.80 & 0.97\\
%     ConvONet ($64 ^3$)  & 4,166,657 & 2.06 & 1.11 \\ 
%     POCO~\cite{boulch2022poco} & 12,790,454 & 5.55  & 62.41 \\
%     \midrule
%     Ours ($3 \times 128^2$) & 2,713,889 & 1.74 & 4.86\\
%     Ours  ($64 ^3$) & 4,778,433 & 2.70 &4.48 \\
%     Ours (w/ our decoder, $3 \times 128^2$) & 2,763,073 & 3.00  & 6.93\\
%     Ours (w/ our decoder, $64 ^3$) & 5,064,801& 3.20 & 6.98\\
%     \bottomrule
%   \end{tabular}}
%  \end{center}
%   \caption{Our method is faster than POCO. Time reported includes mesh generation.}
% \end{table}

The idea of alternating latent topologies could have implications beyond surface reconstruction. Concurrent research has introduced unusual latent topologies, known as \textbf{irregular latents} that show compelling performance benefits for neural fields~\cite{zhang20223dilg}. One can imagine alternating not only between point, triplane, and voxel latents, but also throwing irregular latents in the mix. We are also curious to see if alternating topologies can improve performance on a wide range of tasks in neural fields that require spatially expressive latents, such as semantic or affordance fields.

\section{Acknowledgement}
This project was supported by the US DoD LUCI (Laboratory University Collaboration Initiative) fellowship and partially supported by ARL grants W911NF-20-2-0158 and W911NF-21-2-0104 under the cooperative A2I2 program. D.P. is supported by the Swiss National Science Foundation under grant number P500PT\_206946. G.W. is supported by Samsung, Stanford HAI, and a PECASE from the ARO. L.G. is also supported by an ONR Vannevar Bush Faculty Fellowship. A.K. is also supported by a DARPA Young Faculty Award, NSF CAREER Award, and Army Young Investigator Award.  
%\jj{Finally, discuss how conditional implicit fields generation methods are useful. For example, in "Variational GIGA" or "Synergies Between Affordance and Geometry" they use it for grasping affordance predictions. "Neural Descriptor Fields" models descriptor fields using implicit fields. Some classic applications of conditional implicit fields generation are "PiFu" and 'DiSN'.}

% Our model can reduce the quantization effect due to the grid discretization in the grid form by introducing iterative alternation between grid and point forms. Since point form enables feature learning at individual point level, it is thus capable of encoding more fine-grained local features for conditional occupancy field generation.

%%%%%%%%% REFERENCES
{\small
\bibliographystyle{ieee_fullname}
\bibliography{egbib,bibliography_long,bibliography,bibliography_custom}

\begin{thebibliography}{10}\itemsep=-1pt

\bibitem{Achlioptas2018ICML}
Panos Achlioptas, Olga Diamanti, Ioannis Mitliagkas, and Leonidas~J. Guibas.
\newblock Learning representations and generative models for 3d point clouds.
\newblock In {\em Proc. of the International Conf. on Machine learning (ICML)},
  2018.

\bibitem{Atzmon2020CVPR}
Matan Atzmon and Yaron Lipman.
\newblock {SAL:} sign agnostic learning of shapes from raw data.
\newblock In {\em Proc. IEEE Conf. on Computer Vision and Pattern Recognition
  (CVPR)}, pages 2562--2571, 2020.

\bibitem{boulch2022poco}
Alexandre Boulch and Renaud Marlet.
\newblock Poco: Point convolution for surface reconstruction.
\newblock In {\em Proceedings of the IEEE/CVF Conference on Computer Vision and
  Pattern Recognition}, pages 6302--6314, 2022.

\bibitem{Brock2016ARXIV}
Andr{\'{e}} Brock, Theodore Lim, James~M. Ritchie, and Nick Weston.
\newblock Generative and discriminative voxel modeling with convolutional
  neural networks.
\newblock {\em arXiv.org}, 1608.04236, 2016.

\bibitem{chang2015shapenet}
Angel~X Chang, Thomas Funkhouser, Leonidas Guibas, Pat Hanrahan, Qixing Huang,
  Zimo Li, Silvio Savarese, Manolis Savva, Shuran Song, Hao Su, et~al.
\newblock Shapenet: An information-rich 3d model repository.
\newblock {\em arXiv preprint arXiv:1512.03012}, 2015.

\bibitem{Chen2019NIPS}
Wenzheng Chen, Huan Ling, Jun Gao, Edward Smith, Jaako Lehtinen, Alec Jacobson,
  and Sanja Fidler.
\newblock Learning to predict 3d objects with an interpolation-based
  differentiable renderer.
\newblock In {\em Advances in Neural Information Processing Systems (NIPS)},
  2019.

\bibitem{Chen2019CVPR}
Zhiqin Chen and Hao Zhang.
\newblock Learning implicit fields for generative shape modeling.
\newblock In {\em Proc. IEEE Conf. on Computer Vision and Pattern Recognition
  (CVPR)}, 2019.

\bibitem{Chibane2020CVPR}
Julian Chibane, Thiemo Alldieck, and Gerard Pons-Moll.
\newblock Implicit functions in feature space for 3d shape reconstruction and
  completion.
\newblock In {\em Proc. IEEE Conf. on Computer Vision and Pattern Recognition
  (CVPR)}, 2020.

\bibitem{Choy2016ECCV}
Christopher~Bongsoo Choy, Danfei Xu, JunYoung Gwak, Kevin Chen, and Silvio
  Savarese.
\newblock 3d-r2n2: {A} unified approach for single and multi-view 3d object
  reconstruction.
\newblock In {\em Proc. of the European Conf. on Computer Vision (ECCV)}, 2016.

\bibitem{cciccek20163d}
{\"O}zg{\"u}n {\c{C}}i{\c{c}}ek, Ahmed Abdulkadir, Soeren~S Lienkamp, Thomas
  Brox, and Olaf Ronneberger.
\newblock 3d u-net: learning dense volumetric segmentation from sparse
  annotation.
\newblock In {\em International conference on medical image computing and
  computer-assisted intervention}, pages 424--432. Springer, 2016.

\bibitem{Dai2017CVPR}
Angela Dai, Angel~X. Chang, Manolis Savva, Maciej Halber, Thomas Funkhouser,
  and Matthias Niessner.
\newblock Scannet: Richly-annotated 3d reconstructions of indoor scenes.
\newblock In {\em Proc. IEEE Conf. on Computer Vision and Pattern Recognition
  (CVPR)}, 2017.

\bibitem{dai2020sg}
Angela Dai, Christian Diller, and Matthias Nie{\ss}ner.
\newblock Sg-nn: Sparse generative neural networks for self-supervised scene
  completion of rgb-d scans.
\newblock In {\em Proceedings of the IEEE/CVF Conference on Computer Vision and
  Pattern Recognition}, pages 849--858, 2020.

\bibitem{Deng2020CVPR}
Boyang Deng, Kyle Genova, Soroosh Yazdani, Sofien Bouaziz, Geoffrey Hinton, and
  Andrea Tagliasacchi.
\newblock Cvxnets: Learnable convex decomposition.
\newblock {\em Proc. IEEE Conf. on Computer Vision and Pattern Recognition
  (CVPR)}, 2020.

\bibitem{Deng2020THREEDV}
Zhantao Deng, Jan Bedna{\v{r}}{\'\i}k, Mathieu Salzmann, and Pascal Fua.
\newblock Better patch stitching for parametric surface reconstruction.
\newblock 2020.

\bibitem{Donne2019CVPR}
Simon Donne and Andreas Geiger.
\newblock Learning non-volumetric depth fusion using successive reprojections.
\newblock In {\em Proc. IEEE Conf. on Computer Vision and Pattern Recognition
  (CVPR)}, 2019.

\bibitem{erler2020points2surf}
Philipp Erler, Paul Guerrero, Stefan Ohrhallinger, Niloy~J Mitra, and Michael
  Wimmer.
\newblock Points2surf learning implicit surfaces from point clouds.
\newblock In {\em European Conference on Computer Vision}, pages 108--124.
  Springer, 2020.

\bibitem{Erler2020ECCV}
Philipp Erler, Paul Guerrero, Stefan Ohrhallinger, Niloy~J. Mitra, and Michael
  Wimmer.
\newblock Points2surf learning implicit surfaces from point clouds.
\newblock In {\em Proc. of the European Conf. on Computer Vision (ECCV)}, 2020.

\bibitem{Fan2017CVPR}
Haoqiang Fan, Hao Su, and Leonidas~J. Guibas.
\newblock A point set generation network for 3d object reconstruction from a
  single image.
\newblock {\em Proc. IEEE Conf. on Computer Vision and Pattern Recognition
  (CVPR)}, 2017.

\bibitem{Gadelha2017THREEDV}
Matheus Gadelha, Subhransu Maji, and Rui Wang.
\newblock 3d shape induction from 2d views of multiple objects.
\newblock In {\em Proc. of the International Conf. on 3D Vision (3DV)}, 2017.

\bibitem{Gao2020NIPS}
Jun Gao, Wenzheng Chen, Tommy Xiang, Alec Jacobson, Morgan McGuire, and Sanja
  Fidler.
\newblock Learning deformable tetrahedral meshes for 3d reconstruction.
\newblock In {\em Advances in Neural Information Processing Systems (NIPS)},
  2020.

\bibitem{giebenhain2021air}
Simon Giebenhain and Bastian Goldl{\"u}cke.
\newblock Air-nets: An attention-based framework for locally conditioned
  implicit representations.
\newblock In {\em 2021 International Conference on 3D Vision (3DV)}, pages
  1054--1064. IEEE, 2021.

\bibitem{Gkioxari2019ICCV}
Georgia Gkioxari, Jitendra Malik, and Justin Johnson.
\newblock Mesh {R-CNN}.
\newblock In {\em Proc. of the IEEE International Conf. on Computer Vision
  (ICCV)}, 2019.

\bibitem{Gropp2020ICML}
Amos Gropp, Lior Yariv, Niv Haim, Matan Atzmon, and Yaron Lipman.
\newblock Implicit geometric regularization for learning shapes.
\newblock In {\em Proc. of the International Conf. on Machine learning (ICML)},
  2020.

\bibitem{Groueix2018CVPR}
Thibault Groueix, Matthew Fisher, Vladimir~G. Kim, Bryan~C. Russell, and
  Mathieu Aubry.
\newblock {AtlasNet}: A papier-m\^ach\'e approach to learning 3d surface
  generation.
\newblock In {\em Proc. IEEE Conf. on Computer Vision and Pattern Recognition
  (CVPR)}, 2018.

\bibitem{Haene2017ARXIV}
Christian H{\"{a}}ne, Shubham Tulsiani, and Jitendra Malik.
\newblock Hierarchical surface prediction for 3d object reconstruction.
\newblock {\em arXiv.org}, 1704.00710, 2017.

\bibitem{Hartmann2017ICCV}
Wilfried Hartmann, Silvano Galliani, Michal Havlena, Luc {Van Gool}, and Konrad
  Schindler.
\newblock Learned multi-patch similarity.
\newblock In {\em Proc. of the IEEE International Conf. on Computer Vision
  (ICCV)}, 2017.

\bibitem{Huang2018CVPR}
Xinyu Huang, Xinjing Cheng, Qichuan Geng, Binbin Cao, Dingfu Zhou, Peng Wang,
  Yuanqing Lin, and Ruigang Yang.
\newblock The apolloscape dataset for autonomous driving.
\newblock In {\em Proc. IEEE Conf. on Computer Vision and Pattern Recognition
  (CVPR)}, 2018.

\bibitem{Jiang2020CVPR}
Chiyu Jiang, Avneesh Sud, Ameesh Makadia, Jingwei Huang, Matthias Nie{\ss}ner,
  and Thomas Funkhouser.
\newblock Local implicit grid representations for 3d scenes.
\newblock In {\em Proc. IEEE Conf. on Computer Vision and Pattern Recognition
  (CVPR)}, 2020.

\bibitem{Jiang2018ECCV}
Li Jiang, Shaoshuai Shi, Xiaojuan Qi, and Jiaya Jia.
\newblock {GAL:} geometric adversarial loss for single-view 3d-object
  reconstruction.
\newblock In {\em Proc. of the European Conf. on Computer Vision (ECCV)}, 2018.

\bibitem{jiang2021synergies}
Zhenyu Jiang, Yifeng Zhu, Maxwell Svetlik, Kuan Fang, and Yuke Zhu.
\newblock Synergies between affordance and geometry: 6-dof grasp detection via
  implicit representations.
\newblock {\em arXiv preprint arXiv:2104.01542}, 2021.

\bibitem{Kanazawa2018ECCV}
Angjoo Kanazawa, Shubham Tulsiani, Alexei~A. Efros, and Jitendra Malik.
\newblock Learning category-specific mesh reconstruction from image
  collections.
\newblock In {\em Proc. of the European Conf. on Computer Vision (ECCV)}, 2018.

\bibitem{Kar2017NIPS}
Abhishek Kar, Christian H{\"{a}}ne, and Jitendra Malik.
\newblock Learning a multi-view stereo machine.
\newblock In {\em Advances in Neural Information Processing Systems (NIPS)},
  2017.

\bibitem{kazhdan2013screened}
Michael Kazhdan and Hugues Hoppe.
\newblock Screened poisson surface reconstruction.
\newblock {\em ACM Transactions on Graphics (ToG)}, 32(3):1--13, 2013.

\bibitem{Kazhdan2006PSR}
Michael~M. Kazhdan, Matthew Bolitho, and Hugues Hoppe.
\newblock Poisson surface reconstruction.
\newblock In {\em Proceedings of the Fourth Eurographics Symposium on Geometry
  Processing, Cagliari, Sardinia, Italy, June 26-28, 2006}, volume 256, pages
  61--70, 2006.

\bibitem{kingma2014adam}
Diederik~P Kingma and Jimmy Ba.
\newblock Adam: A method for stochastic optimization.
\newblock {\em arXiv preprint arXiv:1412.6980}, 2014.

\bibitem{Liao2018CVPR}
Yiyi Liao, Simon Donne, and Andreas Geiger.
\newblock Deep marching cubes: Learning explicit surface representations.
\newblock In {\em Proc. IEEE Conf. on Computer Vision and Pattern Recognition
  (CVPR)}, 2018.

\bibitem{lionar2021dynamic}
Stefan Lionar, Daniil Emtsev, Dusan Svilarkovic, and Songyou Peng.
\newblock Dynamic plane convolutional occupancy networks.
\newblock In {\em Proceedings of the IEEE/CVF Winter Conference on Applications
  of Computer Vision}, pages 1829--1838, 2021.

\bibitem{lorensen1987marching}
William~E Lorensen and Harvey~E Cline.
\newblock Marching cubes: A high resolution 3d surface construction algorithm.
\newblock {\em ACM siggraph computer graphics}, 21(4):163--169, 1987.

\bibitem{Ma2021CVPR}
Qianli Ma, Shunsuke Saito, Jinlong Yang, Siyu Tang, and Michael~J. Black.
\newblock {SCALE}: Modeling clothed humans with a surface codec of articulated
  local elements.
\newblock In {\em Proc. IEEE Conf. on Computer Vision and Pattern Recognition
  (CVPR)}, 2021.

\bibitem{Maturana2015IROS}
Daniel Maturana and Sebastian Scherer.
\newblock Voxnet: {A} 3d convolutional neural network for real-time object
  recognition.
\newblock In {\em Proc. IEEE International Conf. on Intelligent Robots and
  Systems (IROS)}, 2015.

\bibitem{mescheder2019occupancy}
Lars Mescheder, Michael Oechsle, Michael Niemeyer, Sebastian Nowozin, and
  Andreas Geiger.
\newblock Occupancy networks: Learning 3d reconstruction in function space.
\newblock In {\em Proceedings of the IEEE/CVF conference on computer vision and
  pattern recognition}, pages 4460--4470, 2019.

\bibitem{Michalkiewicz2019ICCV}
Mateusz Michalkiewicz, Jhony~K Pontes, Dominic Jack, Mahsa Baktashmotlagh, and
  Anders Eriksson.
\newblock Implicit surface representations as layers in neural networks.
\newblock In {\em Proc. of the IEEE International Conf. on Computer Vision
  (ICCV)}, 2019.

\bibitem{Mildenhall2020ECCV}
Ben Mildenhall, Pratul~P Srinivasan, Matthew Tancik, Jonathan~T Barron, Ravi
  Ramamoorthi, and Ren Ng.
\newblock {NeRF}: Representing scenes as neural radiance fields for view
  synthesis.
\newblock In {\em Proc. of the European Conf. on Computer Vision (ECCV)}, 2020.

\bibitem{Niemeyer2020CVPR}
Michael Niemeyer, Lars Mescheder, Michael Oechsle, and Andreas Geiger.
\newblock Differentiable volumetric rendering: Learning implicit 3d
  representations without 3d supervision.
\newblock In {\em Proc. IEEE Conf. on Computer Vision and Pattern Recognition
  (CVPR)}, 2020.

\bibitem{Pan2019ICCV}
Junyi Pan, Xiaoguang Han, Weikai Chen, Jiapeng Tang, and Kui Jia.
\newblock Deep mesh reconstruction from single {RGB} images via topology
  modification networks.
\newblock In {\em Proc. of the IEEE International Conf. on Computer Vision
  (ICCV)}, 2019.

\bibitem{Park2019CVPR}
Jeong~Joon Park, Peter Florence, Julian Straub, Richard~A. Newcombe, and Steven
  Lovegrove.
\newblock Deepsdf: Learning continuous signed distance functions for shape
  representation.
\newblock In {\em Proc. IEEE Conf. on Computer Vision and Pattern Recognition
  (CVPR)}, 2019.

\bibitem{Paschalidou2018CVPR}
Despoina Paschalidou, Ali~Osman Ulusoy, Carolin Schmitt, Luc van Gool, and
  Andreas Geiger.
\newblock Raynet: Learning volumetric 3d reconstruction with ray potentials.
\newblock In {\em Proc. IEEE Conf. on Computer Vision and Pattern Recognition
  (CVPR)}, 2018.

\bibitem{paszke2019pytorch}
Adam Paszke, Sam Gross, Francisco Massa, Adam Lerer, James Bradbury, Gregory
  Chanan, Trevor Killeen, Zeming Lin, Natalia Gimelshein, Luca Antiga, et~al.
\newblock Pytorch: An imperative style, high-performance deep learning library.
\newblock {\em Advances in neural information processing systems}, 32, 2019.

\bibitem{peng2020convolutional}
Songyou Peng, Michael Niemeyer, Lars Mescheder, Marc Pollefeys, and Andreas
  Geiger.
\newblock Convolutional occupancy networks.
\newblock In {\em European Conference on Computer Vision}, pages 523--540.
  Springer, 2020.

\bibitem{qi2017pointnet}
Charles~R Qi, Hao Su, Kaichun Mo, and Leonidas~J Guibas.
\newblock Pointnet: Deep learning on point sets for 3d classification and
  segmentation.
\newblock In {\em Proceedings of the IEEE conference on computer vision and
  pattern recognition}, pages 652--660, 2017.

\bibitem{Qi2017NIPS}
Charles~R Qi, Li Yi, Hao Su, and Leonidas~J Guibas.
\newblock Pointnet++: Deep hierarchical feature learning on point sets in a
  metric space.
\newblock In {\em Advances in Neural Information Processing Systems (NIPS)},
  2017.

\bibitem{Rezende2016NIPS}
Danilo~Jimenez Rezende, S.~M.~Ali Eslami, Shakir Mohamed, Peter Battaglia, Max
  Jaderberg, and Nicolas Heess.
\newblock Unsupervised learning of 3d structure from images.
\newblock In {\em Advances in Neural Information Processing Systems (NIPS)},
  2016.

\bibitem{Riegler2017THREEDV}
Gernot Riegler, Ali~Osman Ulusoy, Horst Bischof, and Andreas Geiger.
\newblock {OctNetFusion}: Learning depth fusion from data.
\newblock In {\em Proc. of the International Conf. on 3D Vision (3DV)}, 2017.

\bibitem{Riegler2017CVPR}
Gernot Riegler, Ali~Osman Ulusoy, and Andreas Geiger.
\newblock Octnet: Learning deep 3d representations at high resolutions.
\newblock In {\em Proc. IEEE Conf. on Computer Vision and Pattern Recognition
  (CVPR)}, 2017.

\bibitem{ronneberger2015u}
Olaf Ronneberger, Philipp Fischer, and Thomas Brox.
\newblock U-net: Convolutional networks for biomedical image segmentation.
\newblock In {\em International Conference on Medical image computing and
  computer-assisted intervention}, pages 234--241. Springer, 2015.

\bibitem{Saito2019ICCV}
Shunsuke Saito, Zeng Huang, Ryota Natsume, Shigeo Morishima, Angjoo Kanazawa,
  and Hao Li.
\newblock Pifu: Pixel-aligned implicit function for high-resolution clothed
  human digitization.
\newblock In {\em Proc. of the IEEE International Conf. on Computer Vision
  (ICCV)}, 2019.

\bibitem{Saito2020CVPR}
Shunsuke Saito, Tomas Simon, Jason~M. Saragih, and Hanbyul Joo.
\newblock Pifuhd: Multi-level pixel-aligned implicit function for
  high-resolution 3d human digitization.
\newblock In {\em Proc. IEEE Conf. on Computer Vision and Pattern Recognition
  (CVPR)}, 2020.

\bibitem{sitzmann2019siren}
Vincent Sitzmann, Julien~N.P. Martel, Alexander~W. Bergman, David~B. Lindell,
  and Gordon Wetzstein.
\newblock Implicit neural representations with periodic activation functions.
\newblock In {\em Proc. NeurIPS}, 2020.

\bibitem{Sitzmann2019NIPS}
Vincent Sitzmann, Michael Zollh{\"{o}}fer, and Gordon Wetzstein.
\newblock Scene representation networks: Continuous 3d-structure-aware neural
  scene representations.
\newblock In {\em Advances in Neural Information Processing Systems (NIPS)},
  2019.

\bibitem{Stutz2018CVPR}
David Stutz and Andreas Geiger.
\newblock Learning 3d shape completion from laser scan data with weak
  supervision.
\newblock In {\em Proc. IEEE Conf. on Computer Vision and Pattern Recognition
  (CVPR)}, 2018.

\bibitem{Takikawa2021CVPR}
Towaki Takikawa, Joey Litalien, Kangxue Yin, Karsten Kreis, Charles~T. Loop,
  Derek Nowrouzezahrai, Alec Jacobson, Morgan McGuire, and Sanja Fidler.
\newblock Neural geometric level of detail: Real-time rendering with implicit
  3d shapes.
\newblock In {\em Proc. IEEE Conf. on Computer Vision and Pattern Recognition
  (CVPR)}, 2021.

\bibitem{tang2021sa}
Jiapeng Tang, Jiabao Lei, Dan Xu, Feiying Ma, Kui Jia, and Lei Zhang.
\newblock Sa-convonet: Sign-agnostic optimization of convolutional occupancy
  networks.
\newblock In {\em Proceedings of the IEEE/CVF International Conference on
  Computer Vision}, pages 6504--6513, 2021.

\bibitem{Tatarchenko2017ICCV}
M. Tatarchenko, A. Dosovitskiy, and T. Brox.
\newblock Octree generating networks: Efficient convolutional architectures for
  high-resolution 3d outputs.
\newblock In {\em Proc. of the IEEE International Conf. on Computer Vision
  (ICCV)}, 2017.

\bibitem{tatarchenko2019single}
Maxim Tatarchenko, Stephan~R Richter, Ren{\'e} Ranftl, Zhuwen Li, Vladlen
  Koltun, and Thomas Brox.
\newblock What do single-view 3d reconstruction networks learn?
\newblock In {\em Proceedings of the IEEE/CVF conference on computer vision and
  pattern recognition}, pages 3405--3414, 2019.

\bibitem{Thomas2019ICCV}
Hugues Thomas, Charles~R. Qi, Jean{-}Emmanuel Deschaud, Beatriz Marcotegui,
  Fran{\c{c}}ois Goulette, and Leonidas~J. Guibas.
\newblock Kpconv: Flexible and deformable convolution for point clouds.
\newblock In {\em Proc. of the IEEE International Conf. on Computer Vision
  (ICCV)}, 2019.

\bibitem{vora2021nesf}
Suhani Vora, Noha Radwan, Klaus Greff, Henning Meyer, Kyle Genova, Mehdi~SM
  Sajjadi, Etienne Pot, Andrea Tagliasacchi, and Daniel Duckworth.
\newblock Nesf: Neural semantic fields for generalizable semantic segmentation
  of 3d scenes.
\newblock {\em arXiv preprint arXiv:2111.13260}, 2021.

\bibitem{Wang2018ECCV}
Nanyang Wang, Yinda Zhang, Zhuwen Li, Yanwei Fu, Wei Liu, and Yu-Gang Jiang.
\newblock Pixel2mesh: Generating 3d mesh models from single rgb images.
\newblock In {\em Proc. of the European Conf. on Computer Vision (ECCV)}, 2018.

\bibitem{williams2022neural}
Francis Williams, Zan Gojcic, Sameh Khamis, Denis Zorin, Joan Bruna, Sanja
  Fidler, and Or Litany.
\newblock Neural fields as learnable kernels for 3d reconstruction.
\newblock In {\em Proceedings of the IEEE/CVF Conference on Computer Vision and
  Pattern Recognition}, pages 18500--18510, 2022.

\bibitem{Wu2015NIPS}
Jiajun Wu, Ilker Yildirim, Joseph~J. Lim, Bill Freeman, and Joshua~B.
  Tenenbaum.
\newblock Galileo: Perceiving physical object properties by integrating a
  physics engine with deep learning.
\newblock In {\em Advances in Neural Information Processing Systems (NIPS)},
  2015.

\bibitem{Wu2016NIPS}
Jiajun Wu, Chengkai Zhang, Tianfan Xue, Bill Freeman, and Josh Tenenbaum.
\newblock Learning a probabilistic latent space of object shapes via 3d
  generative-adversarial modeling.
\newblock In {\em Advances in Neural Information Processing Systems (NIPS)},
  2016.

\bibitem{Xie2019ICCV}
Haozhe Xie, Hongxun Yao, Xiaoshuai Sun, Shangchen Zhou, and Shengping Zhang.
\newblock Pix2vox: Context-aware 3d reconstruction from single and multi-view
  images.
\newblock In {\em Proc. of the IEEE International Conf. on Computer Vision
  (ICCV)}, 2019.

\bibitem{Xu2019NIPS}
Qiangeng Xu, Weiyue Wang, Duygu Ceylan, Radom{\'{\i}}r Mech, and Ulrich
  Neumann.
\newblock {DISN:} deep implicit surface network for high-quality single-view 3d
  reconstruction.
\newblock In {\em Advances in Neural Information Processing Systems (NIPS)},
  2019.

\bibitem{Yang2019ICCV}
Guandao Yang, Xun Huang, Zekun Hao, Ming{-}Yu Liu, Serge~J. Belongie, and
  Bharath Hariharan.
\newblock Pointflow: 3d point cloud generation with continuous normalizing
  flows.
\newblock In {\em Proc. of the IEEE International Conf. on Computer Vision
  (ICCV)}, 2019.

\bibitem{Yao2018ECCV}
Yao Yao, Zixin Luo, Shiwei Li, Tian Fang, and Long Quan.
\newblock Mvsnet: Depth inference for unstructured multi-view stereo.
\newblock {\em Proc. of the European Conf. on Computer Vision (ECCV)}, 2018.

\bibitem{Yao2019CVPR}
Yao Yao, Zixin Luo, Shiwei Li, Tianwei Shen, Tian Fang, and Long Quan.
\newblock Recurrent mvsnet for high-resolution multi-view stereo depth
  inference.
\newblock {\em Proc. IEEE Conf. on Computer Vision and Pattern Recognition
  (CVPR)}, 2019.

\bibitem{Yariv2020ARXIV}
Lior Yariv, Matan Atzmon, and Yaron Lipman.
\newblock Universal differentiable renderer for implicit neural
  representations.
\newblock {\em arXiv.org}, 2003.09852, 2020.

\bibitem{Yariv2020NIPS}
Lior Yariv, Yoni Kasten, Dror Moran, Meirav Galun, Matan Atzmon, Ronen Basri,
  and Yaron Lipman.
\newblock Multiview neural surface reconstruction by disentangling geometry and
  appearance.
\newblock In {\em Advances in Neural Information Processing Systems (NeurIPS)},
  2020.

\bibitem{Yu2020CVPR}
Zehao Yu and Shenghua Gao.
\newblock Fast-mvsnet: Sparse-to-dense multi-view stereo with learned
  propagation and gauss-newton refinement.
\newblock In {\em Proc. IEEE Conf. on Computer Vision and Pattern Recognition
  (CVPR)}, 2020.

\bibitem{zhang20223dilg}
Biao Zhang, Matthias Nie{\ss}ner, and Peter Wonka.
\newblock 3dilg: Irregular latent grids for 3d generative modeling.
\newblock {\em arXiv preprint arXiv:2205.13914}, 2022.

\bibitem{zhao2021point}
Hengshuang Zhao, Li Jiang, Jiaya Jia, Philip~HS Torr, and Vladlen Koltun.
\newblock Point transformer.
\newblock In {\em Proceedings of the IEEE/CVF International Conference on
  Computer Vision}, pages 16259--16268, 2021.

\end{thebibliography}


\begin{thebibliography}{10}\itemsep=-1pt

\bibitem{boulch2022poco}
Alexandre Boulch and Renaud Marlet.
\newblock Poco: Point convolution for surface reconstruction.
\newblock In {\em Proceedings of the IEEE/CVF Conference on Computer Vision and
  Pattern Recognition}, pages 6302--6314, 2022.

\bibitem{cciccek20163d}
{\"O}zg{\"u}n {\c{C}}i{\c{c}}ek, Ahmed Abdulkadir, Soeren~S Lienkamp, Thomas
  Brox, and Olaf Ronneberger.
\newblock 3d u-net: learning dense volumetric segmentation from sparse
  annotation.
\newblock In {\em International conference on medical image computing and
  computer-assisted intervention}, pages 424--432. Springer, 2016.

\bibitem{he2016deep}
Kaiming He, Xiangyu Zhang, Shaoqing Ren, and Jian Sun.
\newblock Deep residual learning for image recognition.
\newblock In {\em Proceedings of the IEEE conference on computer vision and
  pattern recognition}, pages 770--778, 2016.

\bibitem{jiang2021synergies}
Zhenyu Jiang, Yifeng Zhu, Maxwell Svetlik, Kuan Fang, and Yuke Zhu.
\newblock Synergies between affordance and geometry: 6-dof grasp detection via
  implicit representations.
\newblock {\em arXiv preprint arXiv:2104.01542}, 2021.

\bibitem{kazhdan2013screened}
Michael Kazhdan and Hugues Hoppe.
\newblock Screened poisson surface reconstruction.
\newblock {\em ACM Transactions on Graphics (ToG)}, 32(3):1--13, 2013.

\bibitem{lionar2021dynamic}
Stefan Lionar, Daniil Emtsev, Dusan Svilarkovic, and Songyou Peng.
\newblock Dynamic plane convolutional occupancy networks.
\newblock In {\em Proceedings of the IEEE/CVF Winter Conference on Applications
  of Computer Vision}, pages 1829--1838, 2021.

\bibitem{lorensen1987marching}
William~E Lorensen and Harvey~E Cline.
\newblock Marching cubes: A high resolution 3d surface construction algorithm.
\newblock {\em ACM siggraph computer graphics}, 21(4):163--169, 1987.

\bibitem{mescheder2019occupancy}
Lars Mescheder, Michael Oechsle, Michael Niemeyer, Sebastian Nowozin, and
  Andreas Geiger.
\newblock Occupancy networks: Learning 3d reconstruction in function space.
\newblock In {\em Proceedings of the IEEE/CVF conference on computer vision and
  pattern recognition}, pages 4460--4470, 2019.

\bibitem{peng2020convolutional}
Songyou Peng, Michael Niemeyer, Lars Mescheder, Marc Pollefeys, and Andreas
  Geiger.
\newblock Convolutional occupancy networks.
\newblock In {\em European Conference on Computer Vision}, pages 523--540.
  Springer, 2020.

\bibitem{qi2017pointnet}
Charles~R Qi, Hao Su, Kaichun Mo, and Leonidas~J Guibas.
\newblock Pointnet: Deep learning on point sets for 3d classification and
  segmentation.
\newblock In {\em Proceedings of the IEEE conference on computer vision and
  pattern recognition}, pages 652--660, 2017.

\bibitem{ronneberger2015u}
Olaf Ronneberger, Philipp Fischer, and Thomas Brox.
\newblock U-net: Convolutional networks for biomedical image segmentation.
\newblock In {\em International Conference on Medical image computing and
  computer-assisted intervention}, pages 234--241. Springer, 2015.

\bibitem{vora2021nesf}
Suhani Vora, Noha Radwan, Klaus Greff, Henning Meyer, Kyle Genova, Mehdi~SM
  Sajjadi, Etienne Pot, Andrea Tagliasacchi, and Daniel Duckworth.
\newblock Nesf: Neural semantic fields for generalizable semantic segmentation
  of 3d scenes.
\newblock {\em arXiv preprint arXiv:2111.13260}, 2021.

\end{thebibliography}
}

\end{document}

%\title{On Benefits to Majority Groups when Including Data from Minority Groups}

\title{ALTO: Alternating Latent Topologies for Implicit 3D Reconstruction \\ \vspace{0.2cm} Supplemental Material }

\author{Zhen Wang$^1$\thanks{Equal contribution.} \quad Shijie Zhou$^1$\footnotemark[1] \quad  Jeong Joon Park$^2$
\quad Despoina Paschalidou$^2$ \\ Suya You$^3$ \quad  
Gordon Wetzstein$^2$ \quad  Leonidas Guibas$^2$ \quad  Achuta Kadambi$^1$\\
\normalsize{$^1$University of California, Los Angeles \quad
$^2$Stanford University \quad
$^3$DEVCOM Army Research Laboratory}
% {\tt\small {zhenwang, shijiezhou}@ucla.edu}
% \and
% \\
% \\
}
% \and

\maketitle

\section*{Supplementary Content}

\noindent This supplement is organized as follows:
\begin{itemize}[itemsep=0em]
    \item Section~\ref{sec:architecture} contains network architecture details;
    \item Section~\ref{sec:training} contains more details on the training and inference settings;
   \item Section~\ref{sec:ablation} contains more ablation studies of our method;
    \item Section~\ref{sec:shapenet} contains both quantitative and qualitative results on ShapeNet dataset;
    \item Section~\ref{sec:room} contains more qualitative results on Synthetic Room dataset; 
    \item Section~\ref{sec:scannet} contains additional qualitative results on ScanNet dataset;
    \item Section~\ref{sec:code} contains the code link of the comparison baselines; and
    \item Section~\ref{sec:limitation} contains discussion on the limitation of our method and future work. %; and
%    \item Section~\ref{sec:impact} contains discussion on potential negative social impact.
    
\end{itemize}

\section{Network Architecture}
\label{sec:architecture}

\paragraph{PointNet:} Given the input un-oriented point cloud $\mathcal{P}=\left\{\boldsymbol{p}_{i} \in \mathbb{R}^{3}\right\}_{i=1}^{S}$, where $S$ is the number of input points, we map the input coordinates to point features using a fully-connected layer and a ResNet-FC~\cite{he2016deep} block. Instead of using global features as in~\cite{qi2017pointnet}, we use locally-pooled features to fuse local features. 
Specifically, we aggregate features within the same plane or voxel cell from a 2D triplanar or 3D volumetric grids using max-pooling.
We concatenate the locally pooled features with the feature before pooling and then input to the next ResNet block. To obtain the final point features, there are totally 5 ResNet blocks used.

\paragraph{Our ALTO U-Net:} Our alternation U-Net architecture is similar to traditional U-Net~\cite{ronneberger2015u, cciccek20163d}, except that we replace the convolution-only block with our ALTO block where point and grid (either 2D or 3D) features are converted back and forth as depicted in~\fignohref{3} of main paper. The input and output feature dimensions is set to be $32$. There is no ALTO block in the final block of the U-Net. 

\paragraph{Our Attention-based Decoder:} For the triplane representation, we implement 3 single-head attention for 3 feature planes respectively, where the hidden dimension is equal to the feature dimension $32$. For the volume representation, we implement a multi-head attention with $h$ heads. To maximize the flexibility of our method for different datasets and experiments, we set the number of heads $h$ as a hyperparameter and the hidden dimension as $h \times \text{feature dimension} (32)$. The following occupancy network consisting of 5 stacked ResNet-FC blocks with skip connections is used to predict the occupancy probability of query point features. For all experiments, we use a hidden dimension equal to the attention output feature dimension and 5 ResNet blocks for the occupancy network. 

\section{Training and Inference Details}
\label{sec:training}
\paragraph{Object-Level Reconstruction:}
For object-level reconstruction in ShapeNet, we use alternation between latent topologies: point and triplane, because triplane representation is found to tend to give better results for object-level reconstruction in ConvONet~\cite{peng2020convolutional}. The dimension of each 2D feature plane is set as $64^2$. The depth of our ALTO U-Net is $4$, and we do not downsample or upsample in the top two levels of the U-Net, so the lowest resolution of the U-Net is $16^2$.

\paragraph{Scene-Level Reconstruction:}
For scene-level reconstruction, we use alternation between two topologies: point and feature volume. The dimension of the feature volume is set as $64^3$. The depth of our ALTO U-Net is 4, and similarly we do not downsample or upsample in the top two levels, so the lowest resolution of the U-Net is $16^3$. At decoder stage, we set the hyperparameter $h = 4$ for experiments on Synthetic Room dataset and $h=1$ for experiments on ScanNet dataset which we find the best performance in practice.

\paragraph{Mesh Generation:} 

We use a form of Marching Cubes (MC)~\cite{lorensen1987marching} to evaluate occupancy values from implicit representations on a 3D grid.
As a result of Marching Cube, the vertices are usually placed in the middle of segments, which causes discretization effects~\cite{boulch2022poco}. To deal with this issue, we apply the refinement method from POCO~\cite{boulch2022poco}, which takes both the generated vertices and their floor to predict their occupancy values again. After that, we compare two values, mask out non-perfect vertices, take the average between the generated vertices and their floor, and repeat 10 times to improve the granularity. For object-level reconstruction, we use resolution $128$ and for scene-level reconstruction, we use resolution $256$ for marching cubes. 

\paragraph{Hardware:}
We describe the detailed setups that have been used for inference evaluation:
\begin{itemize}
    \item CUDA version: 11.1
    % \item cuDNN version:
    \item PyTorch version: 1.9.0
    \item GPU: single NVIDIA GeForce RTX 3090
    \item CPU: AMD RYZEN PRO 3955WX 16-Cores CPU 
\end{itemize}
\begin{table}[t]

    % \vskip 0.05in
  \begin{center}
  \resizebox{0.55\linewidth}{!}{
  \begin{tabular}{cccccc} 
    \toprule
    Total $\#$ of alternation blocks & IoU $\uparrow$ & Chamfer-$L_1 \downarrow$ & NC$\uparrow$ & F-score$\uparrow$  \\
    \midrule
   0  &  0.831 & 0.55 & 0.912 & 0.892 \\
   3  & 0.847 & 0.50 & 0.914 & 0.910  \\
   6 &  \textbf{0.863} & \textbf{0.47} & \textbf{0.922} & \textbf{0.924} \\
%%% 1head hidden 32 (1105 poco_mc)
%Ours (w/ our decoder, $64^3$) & \textbf{0.912} & \textbf{0.35} & \textbf{0.921} & \textbf{0.981}\\

    \bottomrule
  \end{tabular}}\vspace{-3mm}
 \end{center}
  \caption{\textbf{Ablation study of total number of ALTO  alternation blocks on ShapeNet dataset with 300 input points.}}
   \label{tab:alternation}
\end{table}

\begin{table}[t]

    % \vskip 0.05in
  \begin{center}
  \resizebox{0.55\linewidth}{!}{
  \begin{tabular}{lccccccc} 
    \toprule
  
    Method & IoU $\uparrow$ & Chamfer-$L_1 \downarrow$ & NC$\uparrow$ & F-score$\uparrow$  \\
    \midrule
  ConvONet ($ 3\times128^2$)~\cite{peng2020convolutional} & 0.805 & 0.44 & 0.903 & 0.948
  \\
  ConvONet ($ 64^3$)~\cite{peng2020convolutional} & 0.849 & 0.42 & 0.915 & 0.964  \\
  \midrule
  ALTO ($ 3\times128^2$, Encoder Only) &  0.834 & 0.43 & 0.906 & 0.960 \\
  ALTO ($ 3\times128^2$) &  0.895 & 0.37 & 0.910 & 0.974 \\
  ALTO ($64^3$, Encoder Only) &  0.903 & 0.36 & 0.920 & 0.978 \\
%%% 4heads hidden 128
ALTO ($64^3$) & \textbf{0.914} & \textbf{0.35} & \textbf{0.921} & \textbf{0.981}\\
%%% 1head hidden 32 (1105 poco_mc)
%Ours (w/ our decoder, $64^3$) & \textbf{0.912} & \textbf{0.35} & \textbf{0.921} & \textbf{0.981}\\

    \bottomrule
  \end{tabular}}\vspace{-3mm}
 \end{center}
  \caption{\textbf{Ablation study of our attention-based decoder for different latent topologies used (i.e. point-triplane and point-voxel alternations) on Synthetic Room dataset.} Input points 10K with noise added. Boldface font represents the preferred results.}
   \label{tab:room}
\end{table}
\section{Ablation Studies}
\label{sec:ablation}

In~\cref{tab:alternation}, we report the performance of method with different number of alternations between point and grid forms within each block in the ALTO U-Net. $0$ represents no point-grid alternations (i.e. staying with only grid form), $3$ represents that there is only point-grid alternation in the top two levels of our ALTO U-Net, and $6$ represents that there is point-grid alternations in each level of our ALTO U-Net. As we can see the results, we can observe the trend that increasing the number of ALTO blocks improves the results for all the metrics. 

We also report the results of the ablation study of our attention-based decoder on synthetic room dataset in~\cref{tab:room}. As demonstrated in the table, with our attention-based decoder, it improves results for both triplanar ($3 \times 128^3$) and volumetric representations ($64^3$).

\section{Additional Results on ShapeNet}
\label{sec:shapenet}

\subsection{Quantitative results}
We show per-category quantitative results in ShapeNet with various point density levels: 3K input points (\cref{tab:shapenet3k}), 1K input points (\cref{tab:shapenet1K}) and 300 input points (\cref{tab:shapenet300}). It is notable that when point clouds get sparser, ALTO performs better than POCO on all four metrics for all categories.

\begin{table*}[t]
    % \vskip 0.05in
  \begin{center}
  \resizebox{0.82\linewidth}{!}{
  \begin{tabular}{lcccccccccccc} 
    \toprule
       \multirow{1}{*}{} & \multicolumn{4}{c}{\centering IoU $\uparrow$} & \multicolumn{4}{c}{\centering Chamfer-$L_1$ $\downarrow$} \\
    \cmidrule(l){2-5} \cmidrule(l){6-9} \cmidrule(l){10-13} 
    Method & ONet~\cite{mescheder2019occupancy} & ConvONet~\cite{peng2020convolutional} & POCO~\cite{boulch2022poco} & ALTO & ONet~\cite{mescheder2019occupancy} & ConvONet~\cite{peng2020convolutional} & POCO~\cite{boulch2022poco} & ALTO \\
    \midrule
    Airplane & 0.734 & 0.849 & 0.902 & \textbf{0.908}  &  0.64 & 0.34 & 0.23 & \textbf{0.22} \\
    Bench &  0.682 & 0.830 & 0.865 & \textbf{0.890} & 0.67 & 0.35 & 0.28 & \textbf{0.26} \\
    Cabinet  & 0.855 & 0.940 & 0.960 & \textbf{0.965} & 0.82 &  0.46 & 0.37 & \textbf{0.34} \\ 
    Car  & 0.830 & 0.886 & 0.921 & \textbf{0.924}  &  1.04 & 0.75 & \textbf{0.41} & 0.43 \\
    Chair  &  0.720 & 0.871 & 0.919 & \textbf{0.925}  &  0.95 & 0.46 & 0.33 & \textbf{0.32} \\
    Display  &0.799 & 0.927 & 0.956 & \textbf{0.962}  &  0.82 & 0.36 & 0.28 & \textbf{0.27} \\ 
    Lamp  & 0.546 & 0.785 & \textbf{0.877} & 0.868 &  1.59 & 0.59 & \textbf{0.33} &0.34\\
    Loudspeaker  & 0.826 & 0.918 & \textbf{0.957} & 0.953  &  1.18 & 0.64 & \textbf{0.41} & \textbf{0.41} \\
    Rifle  & 0.668 & 0.846 & 0.897 & \textbf{0.898} & 0.66 & 0.28 &\textbf{0.19} &\textbf{0.19}\\
    Sofa  & 0.865 & 0.936 & 0.963 & \textbf{0.966}  & 0.73 & 0.42 & 0.30 & \textbf{0.29}\\
    Table  & 0.739 & 0.888  & 0.924 & \textbf{0.937}  & 0.76 & 0.38 & 0.31 & \textbf{0.29}\\
    Telephone  &  0.896 & 0.955 & 0.968 & \textbf{0.977} & 0.46 & 0.27 & 0.22 & \textbf{0.21} \\
    Vessel  &  0.729 & 0.865 & \textbf{0.927} & 0.924 & 0.94 & 0.43 & \textbf{0.25} & 0.26\\
    mean  &  0.761 &  0.884 & 0.926 & \textbf{0.931} & 0.87 & 0.44 & \textbf{0.30} & \textbf{0.30} \\
    \midrule
      \multirow{1}{*}{} & \multicolumn{4}{c}{\centering NC $\uparrow$} & \multicolumn{4}{c}{\centering F-score $\uparrow$} \\
    \cmidrule(l){2-5} \cmidrule(l){6-9} \cmidrule(l){10-13} 
    Method & ONet~\cite{mescheder2019occupancy} & ConvONet~\cite{peng2020convolutional} & POCO~\cite{boulch2022poco} & ALTO & ONet~\cite{mescheder2019occupancy} & ConvONet~\cite{peng2020convolutional} & POCO~\cite{boulch2022poco} & ALTO \\
    \midrule
    Airplane  &  0.886 & 0.931 & 0.944 & \textbf{0.949}  & 0.829 & 0.965 & \textbf{0.994} & 0.992  \\
    Bench  & 0.871 & 0.921 & 0.928 & \textbf{0.941}  & 0.827 & 0.964 & 0.988 & \textbf{0.991} \\
    Cabinet  & 0.913 & 0.956 & 0.961 & \textbf{0.967}  & 0.833 & 0.956 & 0.979 & \textbf{0.982} \\ 
    Car  &  0.874 & 0.893 & 0.894 & \textbf{0.917}  & 0.747 & 0.849  & \textbf{0.946} & 0.940 \\
    Chair  & 0.886 & 0.943 & 0.956 & \textbf{0.959} &  0.730 & 0.939 & \textbf{0.985} & \textbf{0.985} \\
    Display  &  0.926 & 0.968 & 0.975 & \textbf{0.976}  &  0.795 & 0.971 & \textbf{0.994} & 0.993 \\ 
    Lamp  &  0.809 & 0.900 & \textbf{0.929} & 0.924  & 0.581 & 0.892 & \textbf{0.975} & 0.962 \\
    Loudspeaker  & 0.903 & 0.939 & \textbf{0.952}  & 0.951  & 0.727 & 0.892 & \textbf{0.964} & 0.955 \\
    Rifle  & 0.849 & 0.929 & \textbf{0.949} & \textbf{0.949}  &  0.818 & 0.980 & \textbf{0.998} & 0.996 \\
    Sofa  & 0.928 & 0.958 & 0.967 & \textbf{0.971}  & 0.832 & 0.953  & \textbf{0.989} & 0.987 \\
    Table  &  0.917 & 0.959 & 0.966 & \textbf{0.968}  & 0.824 & 0.967 & \textbf{0.991}  & 0.990 \\
    Telephone  & 0.970 & 0.983 & 0.985 & \textbf{0.987}  &  0.930 & 0.989 & \textbf{0.998} & \textbf{0.998} \\
    Vessel  & 0.857 & 0.919 & \textbf{0.940} & \textbf{0.940}  & 0.734 & 0.931 & \textbf{0.989} & 0.982 \\
    mean  & 0.891 &0.938 & 0.950 & \textbf{0.954}  & 0.785 & 0.942 & \textbf{0.984} & 0.981 \\
%   ConvONet\\

%   POCO \\
%   \midrule
%   ALTO (Encoder Only) & \\
%   ALTO &  \\
    \bottomrule
  \end{tabular}}
 \end{center}\vspace{-5mm}
  \caption{\textbf{Performance on ShapeNet with input noisy point cloud 3K.} Boldface font represents the preferred results.}
  \label{tab:shapenet3k}
\end{table*}

\begin{table*}[t]
    % \vskip 0.05in
  \begin{center}
  \resizebox{0.82\linewidth}{!}{
  \begin{tabular}{lcccccccccccc} 
    \toprule
       \multirow{1}{*}{} & \multicolumn{4}{c}{\centering IoU $\uparrow$} & \multicolumn{4}{c}{\centering Chamfer-$L_1$ $\downarrow$} \\
    \cmidrule(l){2-5} \cmidrule(l){6-9} \cmidrule(l){10-13} 
    Method & ONet~\cite{mescheder2019occupancy} & ConvONet~\cite{peng2020convolutional} & POCO~\cite{boulch2022poco} & ALTO & ONet~\cite{mescheder2019occupancy} & ConvONet~\cite{peng2020convolutional} & POCO~\cite{boulch2022poco} & ALTO \\
    \midrule
    Airplane & 0.748 & 0.825 & 0.850 & \textbf{0.872}  & 0.59   &  0.39& 0.32&  \textbf{0.29}  \\
    Bench & 0.702 & 0.798 & 0.804 & \textbf{0.856} & 0.62  & 0.40 & 0.38 &  \textbf{0.30}   \\
    Cabinet & 0.862  & 0.926 & 0.936 & \textbf{0.953} & 0.76  & 0.50 & 0.46  & \textbf{0.37} \\
    Car & 0.837 & 0.867 & 0.878 & \textbf{0.901} & 0.99  & 0.83 & 0.60 &  \textbf{0.50} \\
    Chair & 0.736 & 0.837 & 0.867 & \textbf{0.894} & 0.89 & 0.55 & 0.44 & \textbf{0.39} \\
    Display & 0.812 & 0.911 & 0.930& \textbf{0.946} & 0.78 & 0.41  & 0.34 & \textbf{0.31} \\
    Lamp & 0.567 & 0.741 & 0.807 & \textbf{0.820} & 1.44 & 0.68  & \textbf{0.50} & \textbf{0.50} \\
    Loudspeaker & 0.831 & 0.899 & 0.923 & \textbf{0.933} & 1.14 &0.72  & 0.54 & \textbf{0.48} \\
    Rifle & 0.680 & 0.801 & 0.850 & \textbf{0.862} &0.63 & 0.36 & 0.27 & \textbf{0.25}  \\
    Sofa & 0.873& 0.921 & 0.937 & \textbf{0.952}  &0.69 & 0.47 & 0.38 & \textbf{0.33}  \\
    Table & 0.757 & 0.858 & 0.880 & \textbf{0.913} & 0.70& 0.44 & 0.38 & \textbf{0.33}  \\
    Telephone & 0.897 & 0.946 & 0.953 & \textbf{0.968} & 0.46 & 0.29  & 0.26 & \textbf{0.23} \\
    Vessel & 0.736& 0.840 & 0.880 & \textbf{0.893} & 0.91 & 0.51  & 0.37 & \textbf{0.33} \\
    mean & 0.772 & 0.859 & 0.884 & \textbf{0.905} & 0.82 & 0.50  & 0.40 &  \textbf{0.35}\\
    \midrule
      \multirow{1}{*}{} & \multicolumn{4}{c}{\centering NC $\uparrow$} & \multicolumn{4}{c}{\centering F-score $\uparrow$} \\
    \cmidrule(l){2-5} \cmidrule(l){6-9} \cmidrule(l){10-13} 
    Method & ONet~\cite{mescheder2019occupancy} & ConvONet~\cite{peng2020convolutional} & POCO~\cite{boulch2022poco} & ALTO & ONet~\cite{mescheder2019occupancy} & ConvONet~\cite{peng2020convolutional} & POCO~\cite{boulch2022poco} & ALTO \\
    \midrule
    Airplane  & 0.894 & 0.922  & 0.920 & \textbf{0.933} & 0.850 & 0.946 & 0.970 & \textbf{0.976} \\
    Bench & 0.882 & 0.911 & 0.902 & \textbf{0.925} & 0.849 & 0.943 & 0.956 & \textbf{0.979}  \\
    Cabinet & 0.925 &0.949  & 0.945 & \textbf{0.957} & 0.852 & 0.939  & 0.951 & \textbf{0.972} \\ 
    Car & 0.904 & 0.885 & 0.867 & \textbf{0.889} & 0.763 & 0.819 & 0.868 & \textbf{0.912} \\
    Chair & 0.893 & 0.931 &0.930 & \textbf{0.946} & 0.753 &0.902  & 0.943 & \textbf{0.965} \\
    Display & 0.930 & 0.961 & 0.962 & \textbf{0.970} & 0.805 & 0.956 & 0.976 & \textbf{0.984} \\ 
    Lamp & 0.820 & 0.885  & 0.895 & \textbf{0.905} & 0.606 &0.845  & 0.924 & \textbf{0.926}  \\
    Loudspeaker & 0.914 & 0.929 & 0.928 & \textbf{0.936} & 0.740 & 0.863 & 0.908  & \textbf{0.926} \\
    Rifle & 0.859 & 0.916 & 0.928 & \textbf{0.936} & 0.828 & 0.957  & 0.984 & \textbf{0.987} \\
    Sofa & 0.937 & 0.950 & 0.950 & \textbf{0.960} & 0.846 & 0.932 & 0.961 & \textbf{0.974} \\
    Table & 0.918 & 0.950 & 0.949 & \textbf{0.961} & 0.842  & 0.947 & 0.964 & \textbf{0.979} \\
    Telephone & 0.972 & 0.980 & 0.979  & \textbf{0.984} & 0.940 & 0.983 & 0.990 & \textbf{0.994} \\
    Vessel & 0.866  & 0.906 & 0.913& \textbf{0.923} & 0.740 & 0.899 & 0.952  & \textbf{0.961} \\
    mean & 0.901 & 0.929 & 0.928 & \textbf{0.940} & 0.801 & 0.918 & 0.950 & \textbf{0.964} \\
%   ConvONet\\

%   POCO \\
%   \midrule
%   ALTO (Encoder Only) & \\
%   ALTO &  \\
    \bottomrule
  \end{tabular}}
 \end{center}\vspace{-5mm}
  \caption{\textbf{Performance on ShapeNet with input noisy point cloud 1K.} Boldface font represents the preferred results.}
  \label{tab:shapenet1K}
\end{table*}

\begin{table*}[t]
    % \vskip 0.05in
  \begin{center}
  \resizebox{0.82\linewidth}{!}{
  \begin{tabular}{lcccccccccccc} 
    \toprule
       \multirow{1}{*}{} & \multicolumn{4}{c}{\centering IoU $\uparrow$} & \multicolumn{4}{c}{\centering Chamfer-$L_1$ $\downarrow$} \\
    \cmidrule(l){2-5} \cmidrule(l){6-9} \cmidrule(l){10-13} 
    Method & ONet~\cite{mescheder2019occupancy} & ConvONet~\cite{peng2020convolutional} & POCO~\cite{boulch2022poco} & ALTO & ONet~\cite{mescheder2019occupancy} & ConvONet~\cite{peng2020convolutional} & POCO~\cite{boulch2022poco} & ALTO \\
    \midrule
    Airplane & 0.760 & 0.782 & 0.744 & \textbf{0.825} & 0.57 & 0.48 & 0.57 & \textbf{0.39} \\
    Bench & 0.716 & 0.743 & 0.707 & \textbf{0.801} &  0.60 & 0.50 & 0.56 & \textbf{0.39} \\
    Cabinet & 0.867 & 0.900 & 0.889 & \textbf{0.927} &  0.73& 0.52 & 0.58 & \textbf{0.46} \\ 
    Car & 0.834&  0.843  & 0.817  & \textbf{0.867} & 0.99& 0.76 & 0.83 & \textbf{0.67}  \\
    Chair & 0.736 & 0.787  & 0.776 & \textbf{0.840} & 0.89& 0.67 & 0.71 & \textbf{0.52} \\
    Display & 0.817 & 0.885 & 0.878 & \textbf{0.917} & 0.76 & 0.47 & 0.49 & \textbf{0.38} \\ 
    Lamp & 0.567& 0.663 & 0.681 & \textbf{0.747} & 1.38 & 1.02 & 0.93 & \textbf{0.76} \\
    Loudspeaker & 0.827 & 0.870 & 0.867 & \textbf{0.901} & 1.16 & 0.78 & 0.79 & \textbf{0.64} \\
    Rifle & 0.691& 0.757 & 0.742 & \textbf{0.801} & 0.61 & 0.43 & 0.45 & \textbf{0.35}  \\
    Sofa & 0.872 & 0.898  & 0.893 & \textbf{0.926} &  0.69 & 0.52  & 0.53 & \textbf{0.42} \\
    Table & 0.758 & 0.813 & 0.794 & \textbf{0.868} & 0.72 & 0.52 & 0.57& \textbf{0.42} \\
    Telephone & 0.916 & 0.939 & 0.927 & \textbf{0.952} &  0.41 & 0.31 & 0.33 & \textbf{0.27} \\
    Vessel & 0.748 & 0.797 & 0.795 & \textbf{0.846} &  0.85 & 0.63 & 0.60 & \textbf{0.47} \\
    mean & 0.778 & 0.821& 0.808  & \textbf{0.863} &  0.80 & 0.59 & 0.61 & \textbf{0.47} \\
    \midrule
      \multirow{1}{*}{} & \multicolumn{4}{c}{\centering NC $\uparrow$} & \multicolumn{4}{c}{\centering F-score $\uparrow$} \\
    \cmidrule(l){2-5} \cmidrule(l){6-9} \cmidrule(l){10-13} 
    Method & ONet~\cite{mescheder2019occupancy} & ConvONet~\cite{peng2020convolutional} & POCO~\cite{boulch2022poco} & ALTO & ONet~\cite{mescheder2019occupancy} & ConvONet~\cite{peng2020convolutional} & POCO~\cite{boulch2022poco} & ALTO \\
    \midrule
    Airplane & 0.897 & 0.901 & 0.867 & \textbf{0.914} & 0.864 & 0.902 & 0.867 & \textbf{0.938} \\
    Bench & 0.878 &0.886 & 0.864 & \textbf{0.906} & 0.860 & 0.912 & 0.882 & \textbf{0.947} \\
    Cabinet & 0.916& 0.931 & 0.917 & \textbf{0.943} &  0.856 & 0.916 & 0.896 & \textbf{0.943} \\ 
    Car & 0.875&  0.864 & 0.835 & \textbf{0.873} & 0.757 & 0.810 & 0.766 & \textbf{0.850} \\
    Chair & 0.889& 0.905 & 0.885 & \textbf{0.923} & 0.754 & 0.850 & 0.833 & \textbf{0.910} \\
    Display & 0.926 & 0.947 & 0.938 &\textbf{0.956} &  0.813 & 0.926 & 0.916 & \textbf{0.957} \\ 
    Lamp & 0.813 & 0.853 & 0.834 & \textbf{0.875} & 0.618 & 0.771 & 0.781 & \textbf{0.857} \\
    Loudspeaker & 0.897 & 0.911 & 0.897  & \textbf{0.916} & 0.737 & 0.832 & 0.819  & \textbf{0.871} \\
    Rifle & 0.863 & 0.890 & 0.883 & \textbf{0.909} &  0.838 &  0.919  & 0.918 & \textbf{0.952}\\
    Sofa & 0.928 & 0.935 & 0.924 & \textbf{0.946} & 0.846 & 0.906 & 0.899 & \textbf{0.941} \\
    Table & 0.917 & 0.933 & 0.917  &\textbf{0.945}  &  0.839 & 0.913 & 0.894 & \textbf{0.947} \\
    Telephone & 0.970 & 0.975 & 0.970 & \textbf{0.978} & 0.942 & 0.975 & 0.971 & \textbf{0.984} \\
    Vessel & 0.860 & 0.879 & 0.867 & \textbf{0.898} &  0.758 & 0.850 & 0.851 & \textbf{0.909} \\
    mean & 0.895 & 0.908 & 0.892 & \textbf{0.922} &  0.806 & 0.883  & 0.869 &  \textbf{0.924} \\
%   ConvONet\\

%   POCO \\
%   \midrule
%   ALTO (Encoder Only) & \\
%   ALTO &  \\
    \bottomrule
  \end{tabular}}
 \end{center}\vspace{-5mm}
  \caption{\textbf{Performance on ShapeNet with input noisy point cloud 300.} Boldface font represents the preferred results.}
  \label{tab:shapenet300}
\end{table*}

\subsection{Qualitative results}
\begin{figure*}[t]
%   \fbox{\rule{0pt}{6in} \rule{0.9\linewidth}{0pt}}
   \includegraphics[width=\linewidth]{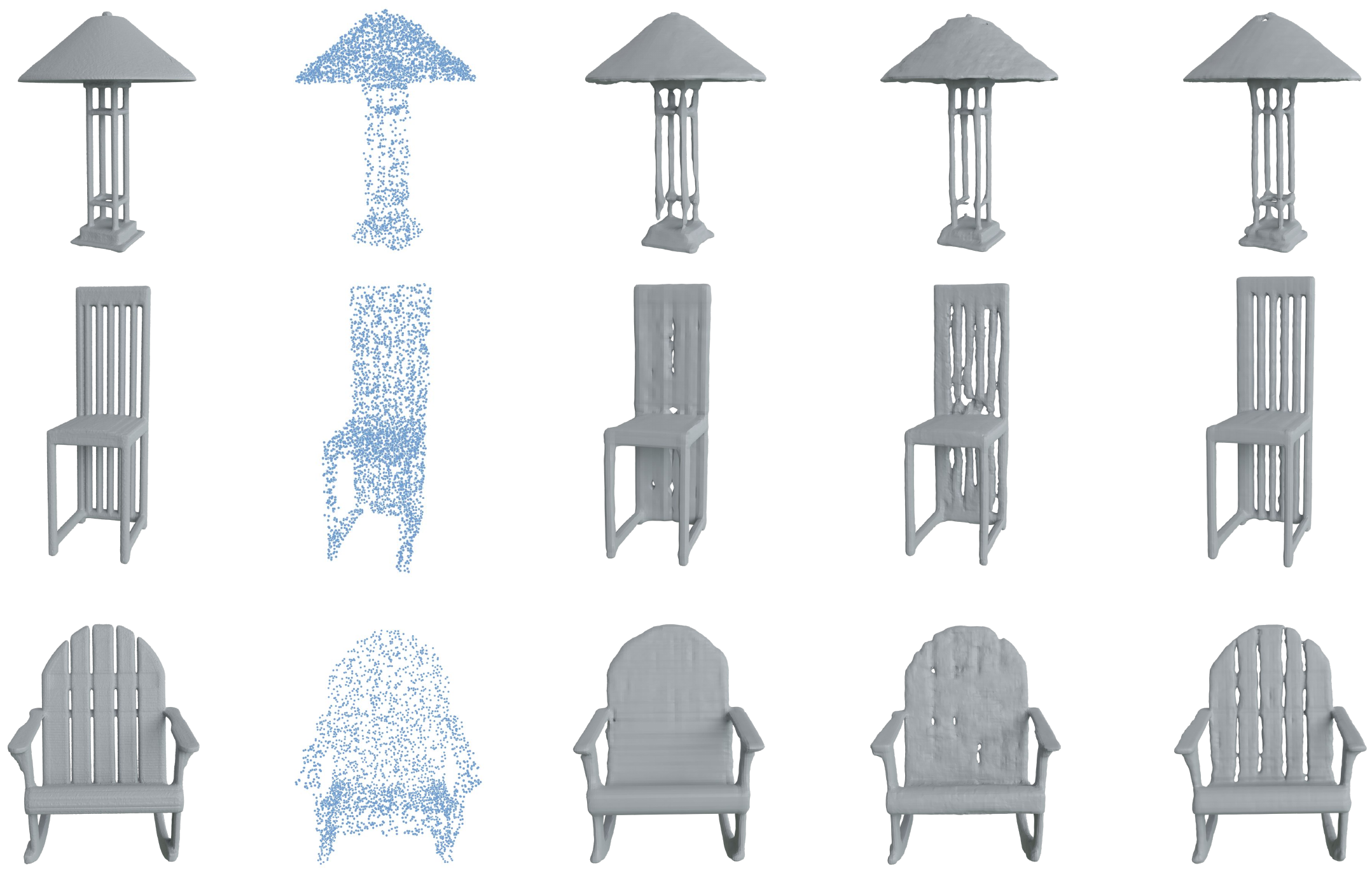}\vspace{-2mm}
\hfill\subcaptionbox{Ground Truth}[0.15\linewidth]
\hfill\subcaptionbox{Input Points}[0.25\linewidth]
\hfill\subcaptionbox{ConvONet~\cite{peng2020convolutional}}[0.18\linewidth]
\hfill\subcaptionbox{POCO~\cite{boulch2022poco}}[0.25\linewidth]
\hfill\subcaptionbox{\textbf{ALTO}}[0.19\linewidth]

\caption{\textbf{Qualitative comparison on object-level reconstruction ShapeNet dataset.} Trained and tested on 3k noisy points.}
\label{fig:shapenet3k}
\end{figure*}

\begin{figure*}[t]
%   \fbox{\rule{0pt}{6in} \rule{0.9\linewidth}{0pt}}
   \includegraphics[width=\linewidth]{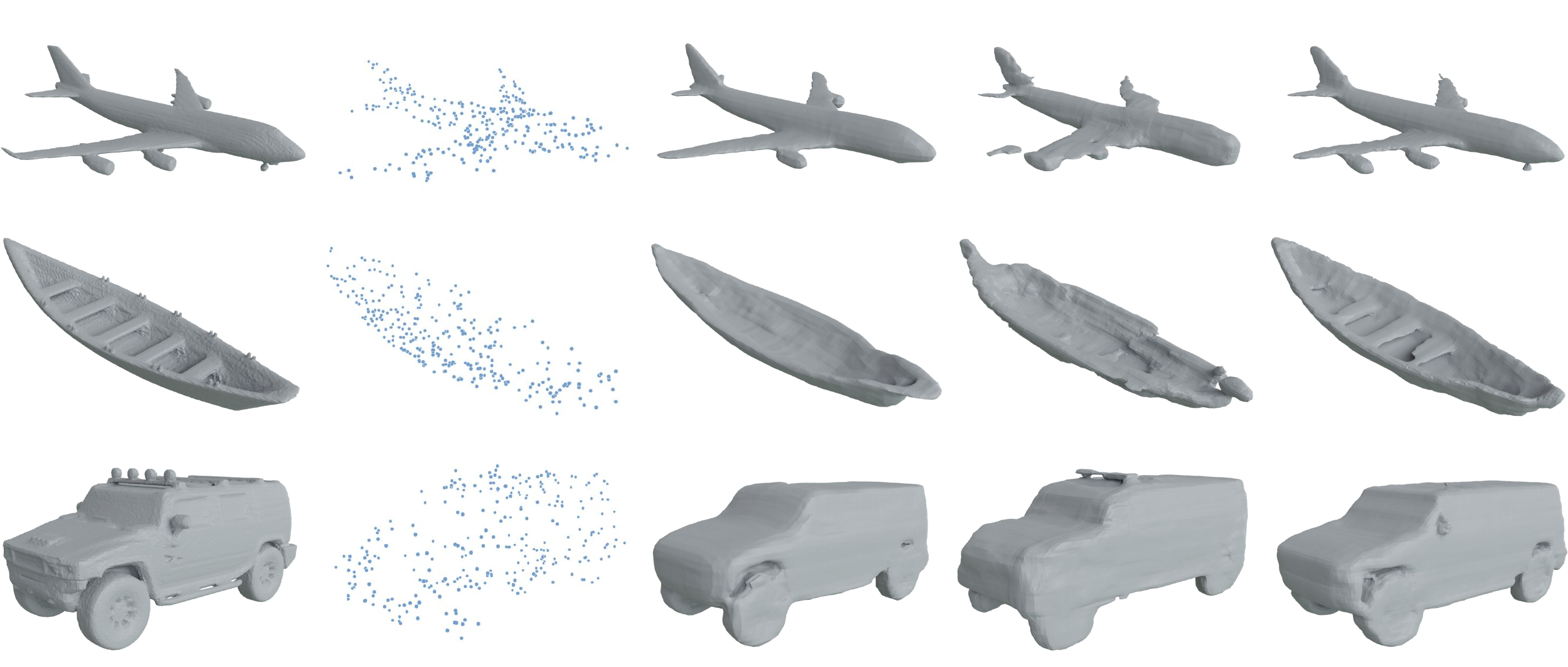}\vspace{-2mm}
\hfill\subcaptionbox{Ground Truth}[0.2\linewidth]
\hfill\subcaptionbox{Input Points}[0.2\linewidth]
\hfill\subcaptionbox{ConvONet~\cite{peng2020convolutional}}[0.2\linewidth]
\hfill\subcaptionbox{POCO~\cite{boulch2022poco}}[0.22\linewidth]
\hfill\subcaptionbox{\textbf{ALTO}}[0.16\linewidth]

\caption{\textbf{Qualitative comparison on object-level reconstruction ShapeNet dataset.} Trained and tested on 300 noisy points.}
\label{fig:shapenet300}
\end{figure*}

Besides 1K input points for ShapeNet as we show in~\fignohref{6} of the main paper, we show additional qualitative results in ShapeNet with 3K input points in~\cref{fig:shapenet3k} and 300 input points in~\cref{fig:shapenet300}.

\section{Additional Results on Synthetic Room Dataset}
\label{sec:room}
\begin{figure*}[t]
%   \fbox{\rule{0pt}{6in} \rule{0.9\linewidth}{0pt}}
   \includegraphics[width=\linewidth]{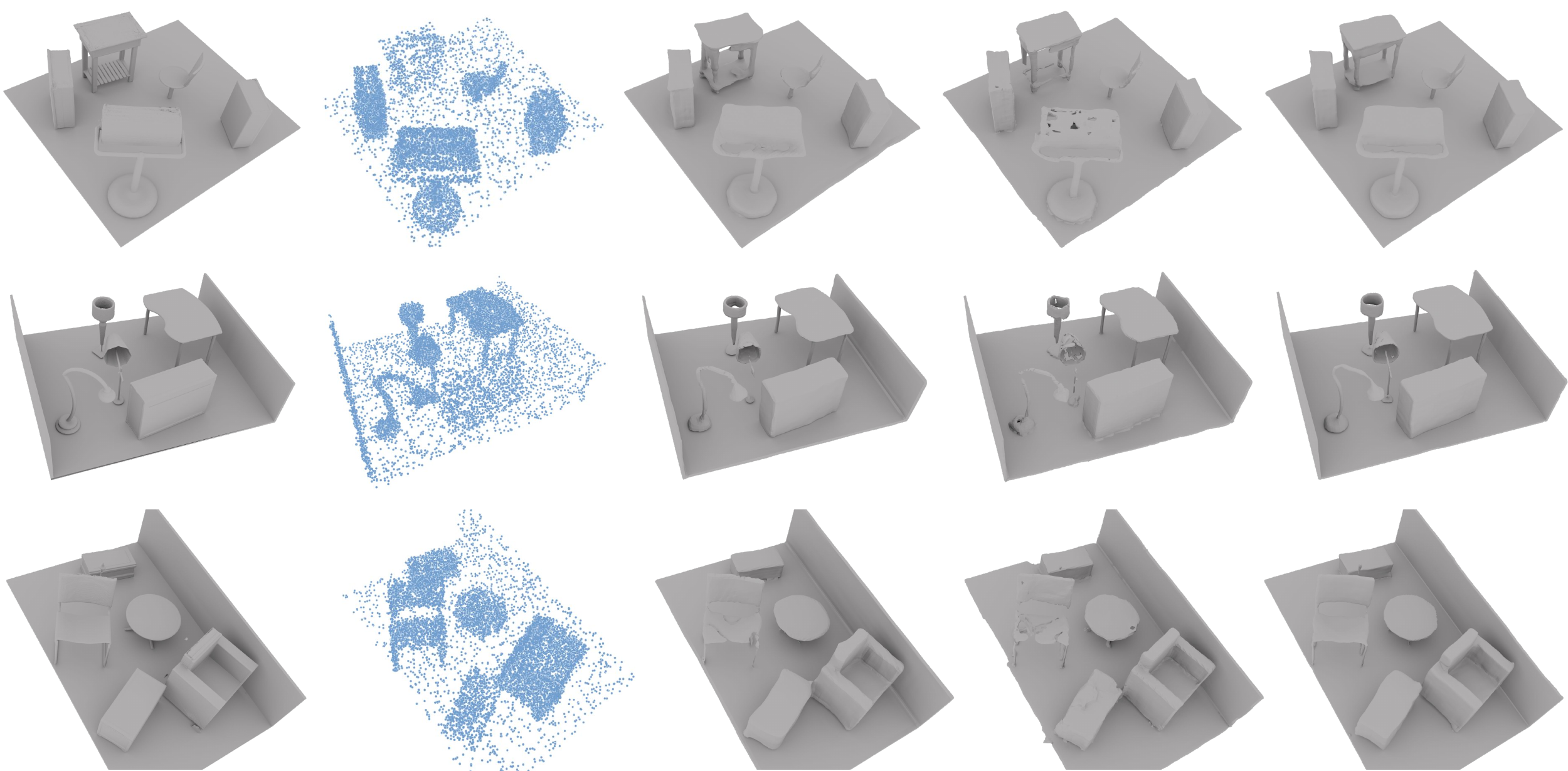}\vspace{-2mm}
\hfill\subcaptionbox{Ground Truth}[0.2\linewidth]
\hfill\subcaptionbox{Input Points}[0.2\linewidth]
\hfill\subcaptionbox{ConvONet~\cite{peng2020convolutional}}[0.2\linewidth]
\hfill\subcaptionbox{POCO~\cite{boulch2022poco}}[0.22\linewidth]
\hfill\subcaptionbox{\textbf{ALTO}}[0.16\linewidth]

\caption{\textbf{Qualitative comparison on scene-level reconstruction Synthetic Room dataset.} Trained and tested on 10k noisy points.}
\label{fig:room10k}
\end{figure*}

\begin{figure*}[t]
%   \fbox{\rule{0pt}{6in} \rule{0.9\linewidth}{0pt}}
   \includegraphics[width=\linewidth]{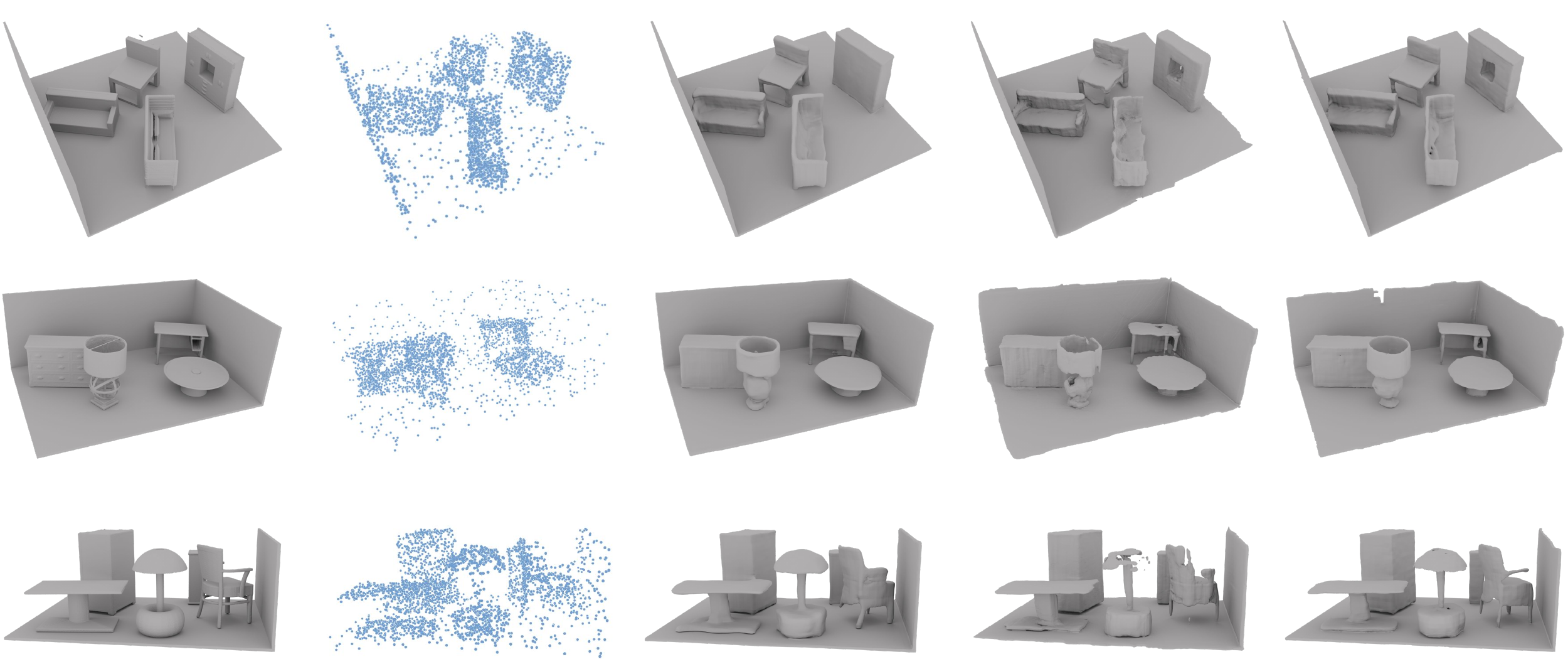}\vspace{-2mm}
\hfill\subcaptionbox{Ground Truth}[0.2\linewidth]
\hfill\subcaptionbox{Input Points}[0.2\linewidth]
\hfill\subcaptionbox{ConvONet~\cite{peng2020convolutional}}[0.2\linewidth]
\hfill\subcaptionbox{POCO~\cite{boulch2022poco}}[0.22\linewidth]
\hfill\subcaptionbox{\textbf{ALTO}}[0.16\linewidth]

\caption{\textbf{Qualitative comparison on scene-level reconstruction Synthetic Room dataset.} Trained and tested on 3K noisy points.}
\label{fig:room3k}
\end{figure*}

We show additional qualitative results in Synthetic Room dataset with 10K inputs points in~\cref{fig:room10k} and 3K inputs points in~\cref{fig:room3k}.

\section{Additional Results on ScanNet}
\label{sec:scannet}
We demonstrate the Sim2Real qualitative results with the model trained on Synthetic Room dataset and tested on ScanNet in \fignohref{8} of the main paper. We show in~\cref{fig:scannet} of the supplement material the Sim2Real results with different point density levels (i.e. $N_{\text{Train}}$=10k, $N_{\text{Test}}$=3k) to further demonstrate the generalization capability of our method ALTO. 

\begin{figure*}[t]
%   \fbox{\rule{0pt}{6in} \rule{0.9\linewidth}{0pt}}
   \includegraphics[width=\linewidth]{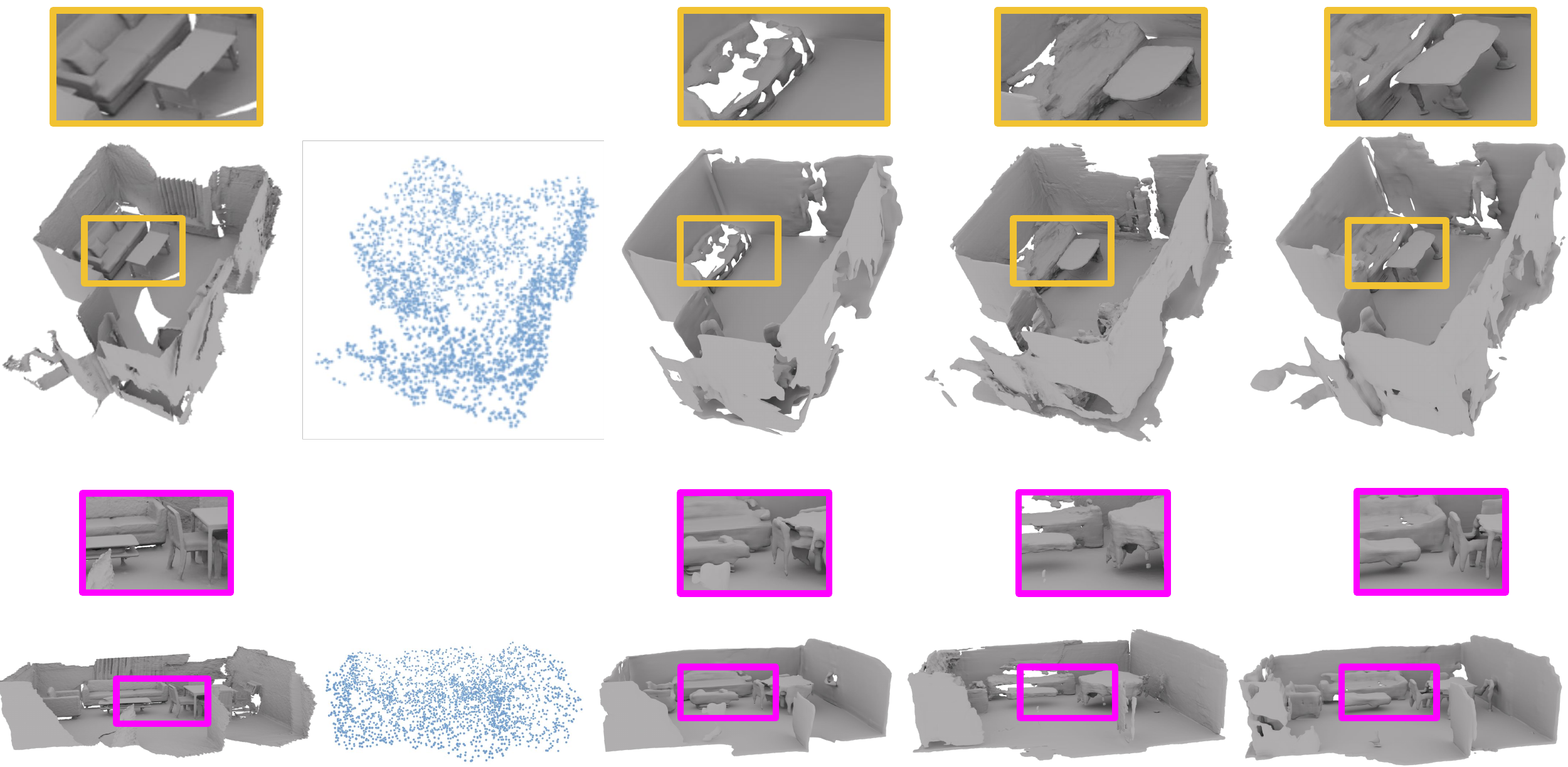}\vspace{-2mm}
\hfill\subcaptionbox{Ground Truth}[0.2\linewidth]
\hfill\subcaptionbox{Input Points}[0.2\linewidth]
\hfill\subcaptionbox{ConvONet~\cite{peng2020convolutional}}[0.2\linewidth]
\hfill\subcaptionbox{POCO~\cite{boulch2022poco}}[0.22\linewidth]
\hfill\subcaptionbox{\textbf{ALTO}}[0.16\linewidth]

\caption{\textbf{Qualitative comparison on scene-level reconstruction ScanNet.}}
\label{fig:scannet}
%   \achuta{specify noise, e.g.g as pct of volume size noise - see previous paper.}
  % \achuta{(a) GT (b) Input .... (f) ours.}
\end{figure*}

\section{Comparison Code Links}
\label{sec:code}

We list all the links of the code of the comparisons baselines in~\cref{tab:comparison_code}. Our code is attached as part of the supplement materials and will be uploaded at \url{https://github.com/cvpr2023-submission/ALTO} upon acceptance.

\begin{table*}[h]
  \centering
  \small
  \begin{tabular}{ll}
    \toprule
    Methods & Links \\
    \midrule
    SPSR~\cite{kazhdan2013screened} & \url{https://github.com/mmolero/pypoisson}\\
    ONet~\cite{mescheder2019occupancy} & \url{https://github.com/autonomousvision/occupancy_networks} \\
    ConvONet~\cite{peng2020convolutional} & \url{https://github.com/autonomousvision/convolutional_occupancy_networks} \\
    DP-ConvONet~\cite{lionar2021dynamic} &  \url{https://github.com/dsvilarkovic/dynamic_plane_convolutional_onet} \\
    POCO~\cite{boulch2022poco} & \url{https://github.com/valeoai/POCO} \\
    \bottomrule
  \end{tabular}
  \caption{\textbf{The link for the baseline methods we compare.}}
  \label{tab:comparison_code} 
\end{table*}

\section{Limitation and Future Work}
\label{sec:limitation}

For our current method, we are not learning a probabilistic generative model that can learn the distribution of the input data, which limits the diversity of the shapes our model can generate. 
Moreover, we are uniformly sampling points as in previous work such as~\cite{peng2020convolutional}. More efficient sampling strategy that 
samples more points on densely populated regions and less on sparsely populated regions can be adopted to capture more details on the fine-grained areas. 

As our method is general in encoding 3D point features, it can be generalized to not just occupancy fields, but also radiance fields trained from images.
Similarly, it can be applied to a broader range of neural fields such as semantic field~\cite{vora2021nesf} and affordance field~\cite{jiang2021synergies}.

%\section{Potential Negative Impact}
%\label{sec:impact}

\clearpage
%%%%%%%%% REFERENCESsa
{\small
\bibliographystyle{ieee_fullname}
\bibliography{egbib_supplement}
}